\documentclass[12pt]{article}

\usepackage{times,epsfig,graphics,amssymb,latexsym,amsmath,setspace,amsthm,fullpage,algorithm}
\usepackage{array,hyperref,url}
\usepackage{float}
\usepackage[round]{natbib}
\usepackage[usenames]{color}

\newtheorem{theorem}{Theorem}

\newtheorem{example}{Example}[section]

\def\argmin{\mathop{\rm argmin}}

\def\sign{\mathop{\rm sign}}

 \newcommand {\bfbeta} {\mbox{\boldmath $\beta$}}
 \newcommand {\bfalpha} {\mbox{\boldmath $\alpha$}}

\def\bv{\mathop{\bf v}}

\def\bx{\mathop{\bf x}}

\def\bK{\mathop{\bf K}}

\def\by{\mathop{\bf y}}

\def\bI{\mathop{\bf I}}

\def\bu{\mathop{\bf u}}
\def\bv{\mathop{\bf v}}

\def\bu{\mathop{\bf u}}

\begin{document}

\pagenumbering{arabic}

\title{\Large \bf  Structure learning via unstructured kernel-based M-regression}
\author{Xin He$^\dag$, Yeheng Ge$^{\dag}$ and Xingdong Feng$^{\dag}$\thanks{Xingdong Feng is the corresponding author.}
	\\[10pt]
	$^\dag$ School of Statistics and Management \\
	Shanghai University of Finance and Economics
}

\date{}
\maketitle

\setcounter{page}{1}
\doublespacing
\begin{abstract}
In statistical learning, identifying underlying structures of  true target functions based on  observed data  plays  a crucial role to facilitate subsequent  modeling and analysis. Unlike most of those existing methods that focus on some specific settings under certain model assumptions, this paper proposes a general and novel framework for recovering true structures of target functions by using unstructured  M-regression in a reproducing kernel Hilbert space (RKHS).  The proposed framework is inspired by the fact that gradient functions can be employed as a valid tool to learn  underlying structures, including sparse learning, interaction selection and model identification, and it is easy to implement by  taking advantage of the nice properties
of the RKHS.  More importantly, it admits a wide range of loss functions, and thus includes many commonly used methods, such as mean regression, quantile regression, likelihood-based classification, and margin-based classification,  which is also computationally efficient by solving convex optimization tasks.
The asymptotic results of the proposed framework are established within a rich family
 of loss functions  without any explicit model specifications.  The superior performance of the proposed framework is also demonstrated by a variety of simulated examples and a real case study.

\end{abstract}

\doublespacing
\section{Introduction}

In statistical learning,  true target functions  are often assumed to have some specific structures to facilitate the following statistical modeling and analysis. Thus,  tremendous interests have been paid to recover underlying structures  from observed data, including learning sparse structures \citep{LiR2012, Hexm2013, WangX2016, Dasgupta2018, HanX2018, Pan2019Ball, Qu2021}, interaction effects \citep{Lin2006, Radchenko2010, Hao2014, Kong2017, Hao2018} or identifying linear and nonlinear effects \citep{ZhangH2011, LianH2015, HEX2020}.
However, most existing methods are designed for learning some specific structures, and their successes   either reply on restrictive model assumptions or require intensive computational efforts. 
For example, various attempts have been made to learn  the sparsity of the conditional mean function by regularization \citep{FAN2010}, screening \citep{FAN2008, WangX2016}, or checking variable robustness against added noises \citep{Barber2015}. The counterparts of these methods have also been proposed in the context of quantile regression  \citep{WuY2009, MaS2017}, margin-based classification   \citep{ Steinwart20082, ZhangWU2016},  generalized linear models \citep{LiY2018} and so on. Furthermore, the additive model assumption is often imposed to relax the linear assumption in pursuing sparsity  \citep{ Huang2010, Fan2011, Lv2018}. However, all these methods  are only designed for some specific learning tasks and lack of universality. Most recently, tremendous attentions have been paid to tackle the universality issue.  \cite{Loh2017} focuses on the theoretical aspect of regularized linear M-estimators within a family of robust loss functions.  \cite{Dasgupta2018}  propose a recursive feature elimination method via repeatedly fitting a kernel ridge regression for a general loss function, but the computational efficiency becomes its main obstacle.  \cite{HanX2018}  proposes a novel nonparametric screening method under a strictly convex loss function family, but it requires  the loss function to be differentiable almost everywhere, which  excludes some popular loss functions, such as the hinge loss.  Other popularly assumed structures of the true target function include the interaction structure \citep{Lin2006, Radchenko2010,Hao2018, Hao2014, Kong2017, Dong2021} and the model identification by identifying linear or nonlinear effects \citep{ZhangH2011, LianH2015, HEX2020}, and these methods are developed in a similar way to the sparse learning. However, these methods are also designed for some specific scenarios, and the lack of theoretical consistency or computational efficiency becomes their main obstacle.

Recently,  many kernel-based sparse  learning methods have been motivated by the fact that  gradient functions provide an appropriate tool to identify informative variables in a model-free fashion, and thus various strategies have been adopted to learn the gradient functions under some specific scenarios.   For example, \cite{Rosasco2013} add an empirical functional penalty on the gradients in a standard kernel ridge regression, and   \cite{HEX20202} further extend it to learn the sparse structure in support vector machines.  \cite{YangL2016} employ a pair-wise learning task to estimate gradient functions and use a functional group lasso penalty to induce sparsity, and  \cite{HEX2020}  extend it to learn interaction structures. Most recently,  \cite{He20192}  propose  an efficient two-step sparse learning method in the least square regression.  Clearly, all these sparse learning methods are methodologically flexible in the sense that they  rely on no model specifications, and thus are applicable to datasets with complicated dependence patterns. However, these methods are developed under specific scenarios in regression or classification, and their high computational costs or lack of  consistency and universality are still unsolved issues.


This paper  proposes a novel  structure learning framework  via  the regularized M-regression for a general family of loss functions in a flexible RKHS. The proposed framework is inspired by the fact that gradient functions characterize structures of their corresponding  true target functions without explicit model assumptions, and the derivative reproducing  properties of the RKHS \citep{ZhouDX2007} facilitate the computation of  gradient functions. The proposed framework is methodologically simple, and computationally easy to implement, which can also be scaled up by a parallelization procedure.  Specifically, it consists of  estimation of the regularized M-regression in a  RKHS, and computation of the gradient functions by a matrix multiplication. It is computationally efficient by only fitting a standard kernel ridge regression via solving a convex optimization problem, and thus scalable to analyze large-scale datasets. More importantly, the asymptotic properties of the proposal in sparse learning, interactions selection and model identification are established based on a general family of loss functions  without imposing any explicit model assumptions.  

The major contributions of the proposed framework are four-fold.
\begin{enumerate}
\item[(i)] It works for  a  general loss function family including most commonly used ones in literature, such as the squared loss, check loss, hinge loss, Huber loss,  logistic loss, $\epsilon$-insensitive loss, exponential loss and so on.
\item[(ii)]  It establishes a unified framework for learning underlying structures of true target functions   and admits general dependence  structures.  The proposed framework  employs gradient functions to recover the structures in a model-free fashion and can be regarded as a joint screening method and thus is able to identify all the informative variables acting on the response with a  general dependence structure, including those marginally noninformative but jointly informative ones.
\item[(iii)] It  is methodologically simple and computationally easy to implement.   Specifically, it avoids directly estimating  gradient functions,  but solving a kernel-based convex optimization problem. Then, the estimated gradient functions  can be efficiently obtained by using the derivative reproducing properties of the RKHS, which  significantly reduces the computational costs.  For instance, in  Examples 1 and 2 of our simulation study, the proposed framework is  efficiently implemented to sparse learning with dimensionality up to $10^5$.
\item[(iv)] It provides  theoretical guarantees for structure learning under mild conditions. With the help of empirical process and functional operators in  learning theory, the estimation consistency of gradient functions is established for a general loss function family. More importantly, as a direct consequence, the asymptotic consistencies of sparse learning, interaction selection and model identification  are established  without imposing any explicit model specifications.
\end{enumerate}

The rest of this paper is organized as follows. Section \ref{sec:pream} introduces the rich family of loss functions and illustrates  the connections between  gradient functions and the corresponding functional structures. Section \ref{sec:proposed} introduces the motivations and the proposed structure learning framework. All the computational details are provided in Section \ref{sec:compute}. In Section \ref{section4}, the asymptotic theoretical results of the proposed method are given. The simulated examples and a real case study are provided in Section \ref{sec:simulated}. A brief discussion is provided in Section \ref{sec:dis}, and extra numerical results and all the technical proofs of Theorems \ref{thm1}--\ref{thm:iden} are deferred to
the Supplementary Material. An R package implementing the proposed method is available at \url{https://github.com/geyh96/GSLM/}.

\section{Preambles and Methodology}\label{sec:pream}

\subsection{A rich family of loss functions}\label{section2}
Suppose a random pair  ${\cal Z}=(\bx,y)$ is drawn from some unknown joint distribution $\rho_{\bx,y}$, with ${\bx}=(x_{1} , ..., x_{p})^T\in {\cal X}$ supported on a compact subset of ${\cal R}^p$ and $y \in {\cal Y}\subset{\cal R}$.  In statistical learning, the true target function  $f^*$ is often defined as the minimizer of the following expected error
\begin{align}\label{Pream:1}
{f^*=\argmin{\cal E}^L(f)=\argmin EL(y, f(\bx)),}
\end{align}
where   ${ L}(\cdot,\cdot): {\cal Y} \times{\cal R}\rightarrow {\cal R}^{+}$ is the loss function of our interests.  We first impose the following conditions on the loss $L$.\\
{\bf Assumption 1:} The loss function ${ L}$ satisfies the following two  conditions.
\begin{itemize}
	\item[(1)] There exist some positive constants $c_1$ and $q\geq 1$ such that ${ L}(y, \omega)\leq c_1 ( |y|^q + |\omega|^q ) $,  for any $y \in {\cal Y}$ and $\omega \in {\cal R}$.
	\item[(2)]  ${ L}(y,\cdot)$ is convex, and locally Lipschitz continuous; that is, for any ${V}\geq 0$, there exists a constant $c_2>0$ such that $\left | { L}(y,\omega) - { L}(y, \omega{'})\right |\leq c_2 |\omega -\omega{'} |$, for any  $\omega, {\omega}{'} \in [-V, V]$ and $y \in {\cal Y}$.
\end{itemize}

Note that the above conditions are mild and commonly used in literature to characterize loss functions \citep{Hang2017, Dasgupta2018}. The loss space satisfying these two conditions include many popular  losses:
\begin{description}
	\item (i) {\bf Squared loss}: ${L}(y,f(\bx))=(y-f(\bx))^2$ with {$c_2=2(M_y+V)$} and $q=2$, for any $|y|\leq M_y$ with a positive constant $M_y$;
	\item (ii) {\bf Check loss}: ${ L}_{\tau}(y,f(\bx))=(y-f(\bx))(\tau-I_{\{y<f(\bx)\}})$ with $ c_2=1$ and $q=1$;
	\item (iii) {\bf Huber loss}: ${ L}(y,f(\bx))=(y-f(\bx))^2$, if $|y-f(\bx)|\leq {\delta}$; $\delta|y-f(\bx)|-\frac{1}{2}\delta^2$, otherwise, with $c_2=\delta$ and $q=1$;
	\item (iv) {\bf $\epsilon$-insensitive loss}: ${ L}(y, f(\bx)) =\max\{0, |y-f(\bx)|-\epsilon\}$ with  $c_2=1$ and $q=1$;
	\item (v) {\bf Logistic loss}: ${ L}(y,f(\bx))= (\ln2)^{-1}\log \big (1+ \exp{(-yf(\bx)}) \big )$ with $c_2=(\ln2)^{-1} e^{V}/(1+e^{V}) $ and $q=1$;
	\item (vi)  {\bf Hinge loss}: ${ L}(y, f(\bx))=(1-yf(\bx))_+$ with  $c_2=1$ and $q=1$;
	\item (vii) {\bf Exponential loss}: ${ L}(y,f(\bx))=  \exp{(-yf(\bx))}$ with $c_2= e^{V} $ and $q=1$.
\end{description}

The explicit form of $f^*$ varies from one loss function to another. For example, when the squared loss is used, $f^*(\bx)=E(y|\bx)$; when the check loss is used,  $f^*(\bx)=Q_{\tau}(y|\bx)$ with $Q_{\tau}(y|\bx)=\inf\{y: P(Y\leq y|\bx)\geq \tau\}$; and when the hinge loss is used,  $f^*(\bx)=\sign(P(y=1|\bx)-1/2)$, where $\sign(\cdot)$ is the sign function. In this paper, we assume that $f^* \in {\cal H}_K$, where ${\cal H}_K$ denotes the RKHS induced by a pre-specified kernel function $K(\cdot,\cdot)$.   This requirement is commonly used in statistical learning literature \citep{Rosasco2013, YangL2016, Dasgupta2018}, and it is well-known that  the RKHS induced by some universal kernels, such as the Gaussian kernel, is a fairly large functional space in that any continuous function can be arbitrarily well approximated by an intermediate function in its induced RKHS under the infinity norm \citep{Steinwart2005}.

\subsection{Structure learning via gradient functions}
In statistical analysis,  the true target function $f^*$ is often assumed to have a specific structure and tremendous interests have been paid to recover the structure of $f^*$ from the observed data, including learning the sparse/interaction structure of $f^*$ or identifying the linear and nonlinear effects in $f^*$.
Unlike most of  existing methods that only work under specific settings and model assumptions, we observe that the gradient functions can be employed as an efficient and flexible tool to meet these statistical interests. Precisely, for the  true target function $f^*$  defined  in   \eqref{Pream:1} with a loss function satisfying Assumption 1, we focus on the first- and second- order gradient functions of $f^*$ that
\begin{align}\label{eq:gradient}
g^*_{l}(\bx)=\frac{\partial f^*(\bx)}{\partial x^l} \ \mbox{and} \ \  g^*_{ lk}(\bx)=\frac{\partial^2f^*(\bx)}{\partial x^l\partial x^{k}},
\end{align}
for $l,k=1,...,p$. In the following, we  illustrate how to use $g^*_{l}(\bx)$ and $g^*_{ lk}(\bx)$ to conduct sparse learning, interaction selection and model identification in the sequential.

\begin{example}\label{example1}
In sparse learning, it is generally believed that only a few  variables have effect on $f^*$, while others are noises \citep{LiR2012, Hexm2013}. By using the first-order gradient function in \eqref{eq:gradient}, we observe  that  a variable $x^l$ does not contribute to  the true target function $f^*$ if and only if
\begin{align}\label{eqn:11}
\| g^*_l \|^2_{{\cal L}^2({\cal X}, \rho_{\bx})} = \int_{\cal X}  ( g_l^*(\bx)  )^2 d\rho_{\bx}=0,
\end{align}
where $\|\cdot \|^2_{{\cal L}^2({\cal X}, \rho_{\bx})}$ denotes  the  ${\cal L}^2({\cal X}, \rho_{\bx})$-norm and $\rho_{\bx}$ denotes the marginal distribution of the covariate  $\bx$. Thus,  evaluating the importance of a variable turns to measure the importance of the corresponding gradient function, and thus $\| g^*_l \|^2_{{\cal L}^2({\cal X}, \rho_{\bx})}$ can be adopted as a valid measure to distinguish the informative and noninformative variables in $f^*$.  Then the true active set can be defined as ${\cal A}^*=\{l: \left \|g^*_l \right \|^2_{{\cal L}^2({\cal X}, \rho_{\bx})} >0 \}$,
\end{example}

\begin{example}\label{example2}
For interaction selection,  many attempts have been made to identify the true interaction effects in underlying models \citep{Lin2006, Radchenko2010, Hao2014, Kong2017,Hao2018, Dong2021}.  We observe that the true interaction effects on $f^*$ can be evaluated by the second-order gradient functions. Specifically, given the true active set ${\cal A}^*$, if a variable $x^l$ has no interaction  effect on $f^*$, the corresponding second-order gradient functions among all the other variables should be zero almost surely in the sense that
\begin{align}\label{eqn:12}
\| g^*_{lk} \|^2_{{\cal L}^2({\cal X}, \rho_{\bx})} = \int_{\cal X}  ( g^*_{lk}(\bx)  )^2 d\rho_{\bx}=0,
\end{align}
for any  $k \in {\cal A}^*$. Thus, the active set containing all the variables that contribute to the  two-way interaction effects in $f^*$ can be defined as
\begin{align*}
{\cal A}^*_{2}=\big \{l \in {\cal A}^*:  \| g^*_{lk} \|^2_{{\cal L}^2({\cal X}, \rho_{\bx})} > 0,~\mbox{for some}\  k \in {\cal A}^* \big \},
\end{align*}
Moreover,  we further denote the set  containing the variables that  only  contribute to the main effects of $f^*$as  ${\cal A}_1^*={\cal A}^* \setminus {\cal A}^*_2$. It is interesting to point out that  the definitions of ${\cal A}^*_1$ and ${\cal A}^*_2$ are general and reduce to those in  \cite{Kong2017} when the true structure of $f^*$ is  a quadratic function.
\end{example}

\begin{example}\label{example3}
Identifying the linear and nonlinear effects in $f^*$ has also attracted many attentions in the literature of partially linear models (PLMs) \citep{ZhangH2011, LianH2015, HEX2020}.  Generally, a PLM considers
\begin{align*}
f^*(\bx)={\bx}^T_{\cal L^*}{\bfbeta}^* + h^*({\bx}_{\cal N^*}),
\end{align*}
where $\bx=({\bx}_{\cal L^*}^T,{\bx}_{\cal N^*}^T)^T \in {\cal R}^p$, ${\cal L^*}^*$ and ${\cal N}^*$ denote the sets of linear and nonlinear effects,  ${\bx}^T_{\cal L}{\bfbeta}^*$ is the linear part and $h^*({\bx}_{\cal N}^*)$ is the nonlinear part.  One of the primal interests is to correctly identify the linear and nonlinear effects in a PLM. We notice that the true linear and nonlinear  effects  can be distinguished by  evaluating the corresponding second-order gradient functions. Specifically, given the true active set ${\cal A}^*$, we observe that if a variable $x_l $ has a linear effect on $f^*$,  the corresponding second-order gradient functions  among all the other variables should be zero almost surely that $\| g^*_{lk} \|^2_{{\cal L}^2({\cal X}, \rho_{\bx})}=0$
for any $k\in {\cal A}^*$. Thus,  the sets of true linear effects  and true nonlinear effects can be defined  as
$
{\cal L}^*=\{ l: \| g^*_{lk} \|^2_{{\cal L}^2({\cal X}, \rho_{\bx})}=0,~\text{for any}~ k \in {\cal A}^* \}
$
and ${\cal N}^*= {\cal A}^*\backslash {\cal L}^*$.
\end{example}

As demonstrated in the above examples, the gradient functions can be employed as an efficient and flexible tool to learn the interested structure of $f^*$ and more importantly,  it provides  appropriate definitions of the interested structure of $f^*$ in a ``model-free'' sense, which avoids the risk of potential model misspecifications. Thus,  it suffices to  learn the corresponding gradient functions consistently and efficiently for identifying the underlying structure of $f^*$.

\section{The Proposed Framework}\label{sec:proposed}

Most existing learning gradient methods formulate  the task into a regularized framework \citep{Rosasco2013, YangL2016, HEX2020,HEX20202} with some carefully designed functional penalties on the gradient functions. However,  these methods usually suffer computational burdens due to the  employed local pair-wise learning tasks or the  added complicated empirical functional penalties. On the contrary, the proposed framework provides an efficient alternative to learning the structure of $f^*$. It is motivated  by the key observations that  the derivative reproducing properties in RKHS \citep{ZhouDX2007} assure that if $ K(\cdot, \cdot) \in { C}^2({\cal X}, {\cal X})$,  then for any $f \in {\cal H}_K$, there holds
\begin{align}\label{rep:pro}
g_l(\bx)=\frac{\partial f(\bx)}{\partial x^l} =\langle f, {\partial_l {K}_{\bx}} \rangle_K, 
\end{align}
where  {${ C}^2$ denotes the class of functions whose second derivative is continuous} and $ {\partial_l {K}_{\bx}(\cdot)}=\frac{\partial K(\bx,\cdot)}{\partial x^l} \in {\cal H}_K$.  Moreover,   if $ K(\cdot, \cdot) \in { C}^4({\cal X}, {\cal X})$, there also holds that
\begin{align}\label{rep:pro2}
g_{lk}(\bx)=\frac{\partial f(\bx)}{\partial x^l \partial x^k } =\langle f, {\partial_{lk} {K}_{\bx}} \rangle_K, 
\end{align}
where $\partial_{lk}K_{\bx} =\frac{\partial K(\bx, \cdot)} {\partial x^l \partial x^k}\in {\cal H}_K$. Note that the facts \eqref{rep:pro} and  \eqref{rep:pro2} assure that to estimate the interested gradient functions within the induced RKHS, it suffices to estimate the target function $f$ itself, and then the  gradient functions can be directly obtained. In the rest of this paper, we focus on the  applications of  the first- and second-order gradient functions to learn the structure of $f^*$ and thus assume that $ K(\cdot, \cdot) \in { C}^4({\cal X}, {\cal X})$, which is naturally satisfied by many kernels, including the Gaussian kernel. Note that it is trivial to extend the proposed framework to estimate arbitrary higher-order gradient functions, which may be useful in some real applications \citep{Ritchie2001}.

Motivated by these key  facts,  we  propose an efficient framework to learn the underlying structure of the true target function $f^*$, which involves a regularized M-estimation in the induced RKHS and the fast computation of  corresponding gradient functions.  Suppose that the random sample ${\cal Z}^n=\{({\bx}_i, y_i)\}_{i=1}^n$ are independent copies of the random pair $(\bx,y)$.  Firstly, we consider the regularized M-estimation in a RKHS to estimate $f^*$ by solving the following optimization problem that
\begin{align}\label{eqn:e3}
\widehat{f}=	\argmin_{f \in {\cal H}_K } \frac{1}{n} \sum_{i=1}^n { L} \left (y_i, f({\bx}_i) \right )   + \lambda \|f\|_K^2,
\end{align}
where the first term is  denoted as ${\cal E}^L_{{\cal Z}^n}(f)$ and $\|\cdot\|_K$ denotes the induced RKHS-norm. By the representer theorem \citep{Wahba1998}, the solution of (\ref{eqn:e3}) must have a finite form that
\begin{align}\label{eqn:22}
\widehat{f}(\bx)=\sum_{i=1}^n\widehat{\alpha}_iK({\bx}_i,\bx)=\widehat{\bfalpha}^T{\bK}_{n}(\bx),
\end{align}
where $\widehat{\bfalpha}=(\widehat{\alpha}_1,...,\widehat{\alpha}_n)^T$ denotes the representer coefficients and ${\bK}_{n}(\bx)=(K({\bx}_1,\bx),...,K({\bx}_n,\bx))^T$ is the $n$-deimensional kernel vector.

Then, we apply the derivative reproducing properties \eqref{rep:pro} and \eqref{rep:pro2} to facilitate the computation of gradient functions of our interests.  Specifically,  once $\widehat{\bfalpha}$ is obtained, we can efficiently compute the estimated first- and second-order gradient functions that
\begin{align*}
&\widehat{g}_l(\bx)=\frac{\partial \widehat{f}(\bx)}{\partial x^l }=\widehat{\bfalpha}^T {\partial_l {\bK}_n({\bx}})\ ~\mbox{and}~ \
\widehat{g}_{lk}(\bx) =\frac{\partial^2 \widehat{f}(\bx)}{\partial x^l \partial x^k}=   \widehat{\bfalpha}^T{\partial_{lk} {\bK}_n({\bx})},
\end{align*}
where $ {\partial_l {\bK}_n({\bx}})=\frac{\partial {\bK}_n({\bx})}{\partial x^l }\in {\cal R}^n$ and $\partial_{lk}\bK_n(\bx)=\frac{\partial^2 {\bK}_n({\bx})}{\partial x^l \partial x^k}\in {\cal R}^n$. Note that once the kernel function $K(\cdot, \cdot)$ is pre-specified, the corresponding gradients $\partial_{l}\bK_n(\bx)$ and $\partial_{lk}\bK_n(\bx)$
 are also analytically determined.

Now we illustrate  how to apply the estimated  gradient functions to recover the underlying structure of the true target function $f^*$ in Examples \ref{example1}--\ref{example3}.   Precisely, for sparse learning,  we  adopt the empirical norm of $\widehat{g}_l$ as a practical measure by computing $ \| \widehat{g}_l \|^2_n= \frac{1}{n} \sum_{i=1}^n \big ( \widehat{g}_l(\bx_i)  \big )^2$, and thus the estimated active set is defined as $\widehat{\cal A}= \left \{l: \left \|\widehat{g}_l\right \|^2_n > v_n \right \}$, where   $v_n$ denotes some pre-specified thresholding value;
for interaction selection, we  adopt the empirical norm of the estimated second-order gradient function by computing $ \| \widehat{g}_{lk}  \|^2_n = \frac{1}{n}\sum_{i=1}^n  \big ( \widehat{g}_{lk}({\bx}_i) \big )^2$, and thus the sets of active interaction and main  effects  in $f^*$ can be estimated as
$$
\widehat{\cal A}_{2}=\big \{l \in\widehat{\cal A}:  \| \widehat{g}_{lk}  \|^2_n >v_n^{int},~\mbox{for some}\ k \in \widehat{\cal A} \big \}  \ \mbox{and} \ \widehat{\cal A}_1= \widehat{\cal A}\setminus \widehat{\cal A}_{2},
$$
respectively,  where $v_n^{int}$ denotes some pre-specified thresholding value. Moreover, we can also apply the estimated second-order gradient functions to conduct model identification, and thus the estimated nonlinear and linear effect sets ${\cal N}^*$ and ${\cal L}^*$ are identified as
$
\widehat{\cal N}=\big \{l \in\widehat{\cal A}:   \| \widehat{g}_{lk} \|^2_n >v_n^{iden},~\mbox{for some}\ k \in \widehat{\cal A} \big \}  \ \mbox{and} \ \widehat{\cal L}= \widehat{\cal A}\setminus \widehat{\cal N},
$
respectively,
where  $v_n^{iden}$ is the pre-defined thresholding value. Note that the structure learning performance of the proposed method highly relies on the choice of pre-specified thresholding values, which can be appropriately determined
through a stability-based selection criterion \citep{SunWW2013} and  more details  are provided in Section \ref{sec:tuning}.

\section{Computational Issues}\label{sec:compute}
In this section, we provide all the computational details as well as the tuning procedures of the proposed framework.

\subsection{Computing algorithms}
Note that the proposed framework is computationally efficient in  that we    only need to solve a convex optimization problem \eqref{eqn:e3}, and then the estimated gradient functions can be directly obtained with the derivative reproducing property of the RHKS. More importantly,
by the representer theorem, the original optimization task  over an infinite function space  ${\cal H}_K$ is converted to an optimization task over a finite $n$-dimensional vector space of $\bfalpha \in {\cal R}^n$.  Specifically, by plugging \eqref{eqn:22}  into \eqref{eqn:e3}, solving the optimization task \eqref{eqn:e3} is equivalent to solving
\begin{align}\label{eq:comp}
\widehat{\bfalpha}=	\argmin_{\bfalpha \in {\cal R}^n } \frac{1}{n} \sum_{i=1}^n { L} \left (y_i, {\bfalpha}^T{\bK}_{n}(\bx)\right )   + \lambda \bfalpha^T \bK \bfalpha,
\end{align}
where $\bK=\{K(\bx_i, \bx_j)\}_{i,j=1}^n\in {\cal R}^{n\times n}$.
Note that  the employed computing algorithm for \eqref{eq:comp} varies from one loss function to another.  For example, for the squared loss function, the solution to \eqref{eq:comp} has an explicit form that $\widehat{\bfalpha}=(\bK^2+n\lambda \bI_n)^{-1}\bK\by$; for the check and hinge loss functions, the dual optimization  can be considered \citep{Takeuchi2006, Boyd2004}, which converts \eqref{eq:comp} to a quadratic programming problem with certain linear constraints; for the logistic loss, the kernel-based  weighted least-square iterations \citep{ZhuJ2005} can be employed. Note that the optimization task \eqref{eq:comp} can be efficiently implemented by using some disciplined convex optimization algorithms, and  the R package \textit{CVXR} \citep{cvxr2020} is used to carry out the  optimization of the proposed framework in all the numerical examples of this paper.

\subsection{Tuning procedure}\label{sec:tuning}
It is interesting to notice that the proposed structure learning framework involves two tuning parameters that the parameter $\lambda$ in \eqref{eq:comp} and the pre-defined thresholding  value used to define active sets of variables. Due to our limited numerical experience, the performance of the proposed framework is satisfactory when $\lambda$ is sufficiently small in various scenarios. Similar observation has also been made in \cite{WangX2016}. Thus, we use $\lambda=10^{-5}$  in Section \ref{sec:simulated}, which yields satisfying performance.

Moreover, we employ the stability-based criterion \citep{SunWW2013}   to select the optimal value of  thresholding parameter. Its key idea is to measure the stability of sparse learning by randomly splitting the training sample into two parts and comparing the disagreement between these two estimated active sets. Specifically, given a thresholding value $v_n$, we randomly split the training sample ${\cal Z}^n$ into two parts ${\cal Z}^n_1$ and ${\cal Z}^n_2$. Then the proposed method is applied to ${\cal Z}^n_1$ and ${\cal Z}^n_2$ and obtain two estimated active sets  $\widehat{\cal A}_{1,v_n}$ and $\widehat{\cal A}_{2,v_n}$, respectively. The disagreement between $\widehat{\cal A}_{1,v_n}$ and $\widehat{\cal A}_{2,v_n}$ is measured by Cohen's kappa coefficient and  the procedure is repeated for multiple times, and then the optimal thresholding value can be determined correspondingly. We refer to \cite{SunWW2013} for more details.

\section{Statistical Properties}\label{section4}
In this section, we provide the  theoretical guarantees of the proposed structure learning framework.  Precisely,  we establish the estimation consistency of gradient functions and provide the asymptotic consistencies of sparse learning, interaction selection and model identification  under mild conditions, respectively.

We start with a brief introduction about some basic knowledge in learning theory. Specifically,  we have $K(\bx,\cdot)\in {\cal H}_{K}$ for any ${\bx} \in {\cal X}$, and $\langle f, {K}_{\bx} \rangle_K=f(\bx)$ for any $f \in {\cal H}_K$. By  Mercer's theorem \citep{Steinwart2008}, under some regularity conditions, the eigen-expansion of the kernel function is given by
\begin{align}\label{eqn:kernel}
K(\bx,\bx)=\sum_{k=1}^{\infty} \mu_k \phi_k(\bx)\phi_k(\bx),
\end{align}
where $\mu_1\geq\mu_2\geq...\geq0$ are  non-negative eigenvalues, and $\{\phi_k\}_{k=1}^{\infty}$ are the associated eigenfunctions, taken to be orthonormal in ${\cal L}^2({\cal X},\rho_{\bx})=\big\{f: \|f\|_2^2<\infty \big\}$.  The RKHS-norm of any $f \in {\cal H}_K$ then can be written as
$$
\|f\|^2_K= \sum_{k\geq 1} \frac{1}{\mu_k} \langle f, \phi_k \rangle^2_{{\cal L}^2({\cal X}, \rho_{\bx})},
$$
which implies that the decay rate of $\mu_k$ fully characterizes the complexity of the RKHS induced by $K$, and has a close relationship with various entropy numbers \citep{Steinwart2008}. Therefore, for any $f \in  {\cal H}_K$, we have $f(\bx)=\sum_{k=1}^{\infty}a_k\phi_k(\bx),$ where $a_k=\langle f, \phi_k\rangle_{{\cal L}^2({\cal X}, \rho_{\bx})}={ \int}_{\cal X}f(\bx)\phi_k(\bx)d\rho_{\bx}$ are Fourier coefficients. Note that these results require that ${\cal H}_K\subset {\cal L}^2({\cal X},\rho_{\bx})$, which is automatically satisfied if $\sup_{\bx \in {\cal X}}K(\bx,\bx)$ is bounded. Moreover, the solution of \eqref{Pream:1} may not be unique, and thus we  further define $f^*=\argmin_{f \in {\cal B} }\|f\|_K^2$ with ${\cal B}=\{f : f= \argmin_{ h \in {\cal H}_K}{\cal E}^L(h)\}$ to ensure the {uniqueness of $f^*$} in the sequel. Furthermore,  we denote  $\widetilde{f}=\argmin_{f \in {\cal H}_{K} }  {\cal E}^L(f) + \lambda\|f\|_K^2$.  We now rewrite  $\lambda$ as  $\lambda_n$ in the rest of this paper, to emphasize its dependency on $n$.

\subsection{Estimation consistency of gradient functions}\label{sec:gradient}
The following technical assumptions are made to establish the estimation consistencies of gradient functions, which is crucial to ensure the asymptotic consistency of the proposed structure learning framework. We further introduce following assumptions.

{\noindent \bf Assumption 2:}  There exist some positive constants $\kappa_1$ and $\kappa_2$  such that $\sup_{\bx \in {\cal X}}\|K_{\bx}\|_K \leq \kappa_1$ and $\sup_{\bx \in {\cal X}} \|\partial_l K_{\bx}\|_K \leq \kappa_2$ for any $l = 1, ..., p$.

{\noindent \bf Assumption 3:}  There exist some positive constants $c_3$ and  $\theta$ such that the approximation error $\|\widetilde{f} -f ^*\|_K=c_3\lambda_n^{\theta}$.

Assumption 2 imposes the boundedness condition on the kernel function as well as the corresponding gradient functions. This assumption is commonly used in machine learning literature  \citep{ Rosasco2013, YangL2016} and satisfied by many kernels with the compact support condition, including the Gaussian kernel, Sobolev kernel, scaled linear kernel, scaled quadratic kernel and so on. Note that the requirement of compact support is usually assumed in machine learning literature for mathematical simplicity, and many efforts have been made to extend it to the non-compact setting \citep{Simon2018}. Assumption 3 quantifies the approximation error as a function of the tuning parameter $\lambda_n$, which is  sensible as $\lim_{\lambda_n\rightarrow 0} \|\widetilde{f}-f^*\|^2_K=0$ in general. Similar assumptions are also used  in literature to control the approximation error rate \citep{ Mendelson2010, Rosasco2013,  ZhangC2016, Dasgupta2018}. Particularly,  \cite{Mendelson2010} prove that the approximation error  under the squared loss function can be explicitly quantified as $O(\lambda_n^{r-1/2})$ with $r \in (1/2, 1]$. Further investigations about the approximation error rate  are provided in  \cite{Eberts2013} by imposing some additional technical assumptions.

\begin{theorem}\label{thm1} Suppose that Assumptions 1--3 are satisfied. Let $\lambda_n=n^{-1/(4q)}$, then for any $\delta_n\geq  2 (\log n)^{-1/q}E|y|$, there exists some positive constant $c_4$ such that with probability at least $1-\delta_n$, such that the following inequality holds
\begin{align}
 \max_{l=1,...p} \Big |\|\widehat{g}_l\|^2_n- \|g^*_l\|^2_{{\cal L}^2({\cal X}, \rho_{\bx})} \Big | \leq  {c_4} \left( \log \frac{4p}{\delta_n} \right)^{1/2} (\log n)^{q/2}n^{-\Theta},
\end{align}
where $\Theta=\min\{\frac{3}{16}, -\frac{\theta}{4q} \}$, $q$ and $\theta$ are given in Assumptions 1 and 3, respectively.
\end{theorem}
Theorem \ref{thm1} establishes the estimation consistency of the estimated first-order gradients $\widehat{g}_l, l=1,...,p$, in the sense that $\|\widehat{g}_l\|^2_n$ converges to $\|g^*_l\|^2_{{\cal L}^2({\cal X}, \rho_{\bx})}$ with high  probability, and is crucial to recovery the underlying structure of $f^*$. Note that this convergence result is established without any model assumption on $f^*$ and holds true for a general loss $L$ satisfying Assumption 1, which includes many scenarios as its special cases, such as mean regression, quantile regression, likelihood-based classification, and margin-based classification. Specially,  for the binary classification, the upper bound  in Theorem \ref{thm1}  reduces to $c_4   \big ( \log \frac{4p}{\delta} \big  )^{1/2} n^{-\Theta}$ for any $\delta\in (0,1)$. It is also worthy pointing out that once the squared loss is used,  the convergence rate in  Theorem \ref{thm1} can be further strengthened to obtain a faster strong convergence rate   \citep{Fischer2020} if some additional technical assumptions, such as the decay rate of $\mu_k$ in (\ref{eqn:kernel}),  are met.

The following technical assumption is made to establish the estimation consistency of the second-order gradient functions.

\noindent{\bf Assumption 4}: There exists some constant $\kappa_3$ such that $\sup_{\bx \in {\cal X}} \|\partial_{lk} K_{\bx}\|_{K}\leq \kappa_3$, for  any $l, k=1,...,p$.

Assumption 4 can be regarded as the extension of Assumptions 2 by requiring
the boundedness of the second-order gradients of $K_{\bx}$, and  is also naturally satisfied by all the kernels  discussed after Assumption 2.

\begin{theorem}\label{inter:thm4}
	Suppose all the assumptions of Theorem \ref{thm1} as well as Assumption 4 are met. Then, there exists some positive constant $c_5$ such that with probability at least $1-\delta_n$, there holds
	$$
 	\max_{l,k =1,...,p} \ \big|  \| \widehat{g}_{lk}   \|^2_n - \| g^*_{lk}\|_{{\cal L}^2({\cal X}, \rho_{\bx})}^2  \big |  \leq  c_5 \Big ( \log \frac{4p^2}{\delta_n} \Big )^{\frac{1}{2}}  (\log n)^{q/2} n^{-\Theta},
	$$
	where $\Theta=\min\{\frac{3}{16}, \frac{\theta}{4q} \}$, $\delta_n$, $q$ and $\theta$ are given in Theorem \ref{thm1}.
\end{theorem}
Theorem \ref{inter:thm4} shows that the estimated second-order gradient function $\left \|\widehat{g}_{lk}\right \|^2_n$ converges to $\left \| g_{lk}^* \right \|_{2}^2$ with high probability, which is crucial to establish the consistency for  the application to  interaction selection and model identification.  It is worthy pointing out that the estimation consistency of arbitrary higher-order gradient functions can  also be established by requiring
the boundedness of  corresponding higher-order gradients of $K_{\bx}$, which is  naturally satisfied by many popularly used kernels, such as the Gaussian kernel.

\subsection{Theoretical property of sparse learning}
In this section, we use the obtained  theoretical results in Section \ref{sec:gradient} to  establish the selection consistency of the proposal in the sparse learning  given in Example \ref{example1} of Section \ref{section2} by using the first-order gradient functions. The following technical assumption is needed to establish the theoretical result.

{\noindent \bf Assumption 5.} There exist some positive constants $c_6$ and $\xi_1>\frac{q}{2}$  such that  $\min_{l\in {\cal          A}^*}\|g^*_{l}\|^2_{{\cal L}^2({\cal X}, \rho_{\bx})}  > c_6\big ( \log \frac{4p}{\delta_n} \big  )^{1/2} (\log n)^{\xi_1}n^{-\Theta}$, where $\Theta$ is given in Theorem \ref{thm1}.

Assumption 5 requires that the true gradient function $g^*_l$ should contain sufficient information about the truly informative variables, and it can be regarded as a condition on the required minimal signal strength, which may go to zero with the increase of sample size. Note that this assumption is crucial to establish the selection consistency  and  is much weaker than many nonparametric sparse learning methods \citep{Huang2010, YangL2016}, which often require the signal is  bounded away from zero by some positive constants. 

\begin{theorem}[Sparse learning]\label{thm:sparse} Suppose  all the assumptions of Theorem \ref{thm1} as well as Assumption 5 are satisfied.  Let $v_n=\frac{c_6}{2}\big ( \log \frac{4p}{\delta_n} \big  )^{1/2} (\log n)^{\xi_1}n^{-\Theta}$, then we have
	\begin{align}
	P(\widehat{\cal A}={\cal A}^*) \rightarrow 1.
	\end{align}
\end{theorem}
Theorem \ref{thm:sparse} shows that the estimated informative set in sparse learning can exactly recover the true active set  with high probability.
 This result is fascinating and attractive given the fact that it holds true for a general loss function satisfying Assumption 1, and thus includes many scenarios as its special cases, such as mean regression, quantile regression, likelihood-based classification, and margin-based classification. Moreover, this result is established without requiring any pre-specified model assumption and allows general dependence structures among variables and response in a model-free fashion.

\subsection{Theoretical guarantees for interaction selection/model identification}
In this section, we use the obtained  theoretical results in Section \ref{sec:gradient} to  establish  the  consistency  of the proposal in interaction selection and model identification by using the second-order  gradient functions  in Theorems \ref{inter:thm5} and \ref{thm:iden}, respectively.

Firstly,  we consider the interaction selection as given in Example \ref{example2} of Section \ref{section2}  and  the following technical assumption is required.

\noindent{\bf Assumption 6}: There exist some positive constants $c_7$ and $\xi_2>\frac{q}{2}$ such that
$\min_{\substack{l,k \in {\cal A}^*_{2}} } \|g^*_{lk}\|_{{\cal L}^2({\cal X}, \rho_{\bx})}^2 >c_7\Big ( \log \frac{4p_0^2}{\delta_n} \Big )^{{1}/{2}}  (\log n)^{\xi_2} n^{-\Theta}$, where $\Theta$ is given in Theorem \ref{thm1}.

Assumption 6 can be regarded as the extension of Assumption 5 by requiring  the true second-order gradient functions have sufficient information about the interaction effects.

\begin{theorem}[Interaction selection consistency]\label{inter:thm5}
	Suppose that the assumptions of Theorem \ref{thm:sparse} as well as Assumption 6 are met. By taking $v^{int}_n=\frac{c_7}{2}\Big ( \log \frac{4p_0^2}{\delta_n} \Big )^{{1}/{2}}  (\log n)^{\xi_2} n^{-\Theta}$, we have
	\begin{align*}
	P \Big ( \widehat{\cal A}_{2}={\cal A}_{2}^*, \widehat{\cal A}_1={\cal A}_1^* \Big )\rightarrow 1, \ \ \mbox{as} \ \ n\rightarrow \infty.
	\end{align*}
\end{theorem}
Theorem \ref{inter:thm5} shows that the proposal used in the interaction selection  can exactly detect all the interaction effects with high probability. Note that this result is established without imposing the strong heredity assumption, which is often assumed by the existing parametric interaction selection methods \citep{Hao2014}. Clearly,  the proposed method can be extended to detect higher-order interaction effects, which is of particular interests in some real applications \citep{Ritchie2001}.  It is also worthy pointing out that the interaction selection consistency is  established  for  a rich loss function family with a  general kernel, which allows detecting general interaction structures among variables for various scenarios.

Finally,  we turn to establish the  consistency of model identification as illustrated in Example \ref{example3} of Section \ref{section2} and  the following technical assumption is introduced.

\noindent{\bf Assumption 7.} There exist some positive constants {$c_8$} and $\xi_3>\frac{q}{2}$  such that
	$\min_{\substack{l,k \in {\cal N}^*} } \|g^*_{lk}\|_{{\cal L}^2({\cal X}, \rho_{\bx})}^2 >c_8\Big ( \log \frac{4p_0^2}{\delta_n} \Big )^{{1}/{2}}  (\log n)^{\xi_2} n^{-\Theta}$, where $\Theta$ is given in Theorem \ref{thm1}.

Assumption 7 requires that the gap of signal strengths between linear and nonlinear effect in the sense  that the corresponding second-order gradient functions of the linear effect are exactly zero, and those of  the nonlinear effect are lower bounded.

\begin{theorem}\label{thm:iden}
	Suppose   that all the assumptions in Theorem  \ref{thm:sparse} as well as Assumption 7 are met. By taking $v^{iden}_n=\frac{c_8}{2}\Big ( \log \frac{4p_0^2}{\delta_n} \Big )^{{1}/{2}}  (\log n)^{\xi_2} n^{-\Theta}$, we have
	\begin{align*}
	P \Big ( \widehat{\cal L}={\cal L}^*, \widehat{\cal N}={\cal N}^* \Big )\rightarrow 1, \ \ \mbox{as} \ \ n\rightarrow \infty.
	\end{align*}
\end{theorem}
Theorem \ref{thm:iden} shows that the underlying model structure can be exactly identified  with probability tending to 1. This theoretical result is also established for a general loss function satisfying Assumption 1  without any explicit model specifications. It provides strong theoretical support for  automatically discovering the model structure for the PLMs, which is particularly attractive in the field of partially linear models.

\noindent{\bf Remark.}  It is worthy pointing out that Theorems  \ref{inter:thm5} and \ref{thm:iden} are established under the case that  noise variables are also included in the collected variable set that $|{\cal A^*}|\ll p$,  and thus the proposal in sparse learning is used to recovery all the informative variables at first, and then  either interaction selection or model identification with the proposed method are conducted based on the  variables identified at the first step. In some other scenarios, where all the collected variables are believed to be related with the response that ${\cal A}^*=\{1,...,p\}$, the proposed framework can be directly applied without applying  sparse learning and the similar theoretical results can be obtained.

\section{Numerical Studies}\label{sec:simulated}
In this section, the proposed framework is applied to sparse learning and interaction selection, and  its numerical performance  are compared with various state-of-the-art competitors under several settings. For the proposed framework,  the RKHS induced by the Gaussian kernel  $K(\bu,\bv)= \exp{\left(-\frac{\|\bu-\bv\|^2}{2\sigma_n^2}\right)} $ is adopted in all the examples, where $\sigma_n$ is set as the median of all the pairwise distances among the training sample. Other tuning parameters such as the thresholding value are selected by the stability-based criterion \citep{SunWW2013} as introduced in Section \ref{sec:tuning} via a grid search, where the grid is set as $\{10^{-3+0.1s} : s = 0, ..., 60\}$.

\subsection{Application to sparse  learning}\label{sec:app1}
In this part, the application of the proposed framework to sparse learning is considered. Specifically,  we consider regression  with  the squared loss, the check loss with $\tau=0.5$ and the Huber loss, and classification with the hinge loss and the logistic loss, due to their popularity and importance in statistical machine learning \citep{ZhuJ2005,  Hexm2013,   YangL2016, ZhangWU2016}, and denoted as GSLM-SQ, GSLM-QA, GSLM-HB, GSLM-SVM and GSLM-LOG for simplicity. Under regression setting,  we consider five competitors, including distance correlation learning (DC, \cite{LiR2012}), the quantile-adaptive screening (QaSIS, \cite{Hexm2013}), the sure independence rank screening (SIRS, \cite{zhu2011SIRS}), the modified Blum-Kiefer-Rossenblatt correlation (MBKR, \cite{zhou2018MBKR}) and the generic sure independence screening (Ball, \cite{Pan2019Ball}).  Under classification setting, we also consider five competitors, including DC, SIRS, MBKR, the screening procedure based on empirical conditional distribution (MV-SIS,  \cite{cui2015MVxy}), and the Kolmogolov Filter (Kol-Filter, \cite{mai2013kolmogorov}). Note that the screening-based competitors are suggested to keep the first $[n/\log n]$ variables to assure the sure screening property, and for fair comparison and saving the space, we here report the results for those competitors truncated by
the thresholding values based on the stability-based criterion as introduced at the beginning of this section to conduct sparse learning, and we denote the truncated versions with the suffix  ``-t'', such as DC-t and QaSIS-t. More numerical results of those competing methods implemented as suggested by their authors are reported in the Supplementary Material.

The following two simulated examples are examined under various scenarios.

{\noindent\bf Example 1} ( Regression):
We first generate $x_i = (x_{i1}, ..., x_{ip})^T$ with $x_{ij} = \frac{W_{ij} +\eta U_i}{1+\eta}$, where $W_{ij}$ and $U_i$ are independently drawn from $U(-0.5, 0.5)$. The response $y_i$ is generated as
$
y_i=8f_1(x_{i1})+4f_2(x_{i2})f_3(x_{i3})+6f_4(x_{i4})+5f_5(x_{i5})+\epsilon_i,
$
where $f_1(u)=u, f_2(u)=2u+1, f_3(u)=2u-1, f_4(u)=0.1\sin(\pi u)+0.2\cos(\pi u)+0.3(\sin(\pi u))^2+0.4(\cos(\pi u))^3+0.5(\sin(\pi u))^3$, $f_5(u)=\sin(\pi u)/(2-\sin(\pi u))$, and $\epsilon_i$'s are independently drawn from $N(0,1)$. Clearly, the first five variables are truly informative.

{\noindent\bf Example 2} (Classification):
We  generate $x_i = (x_{i1}, ..., x_{ip})^T$ with $x_{ij} = \frac{W_{ij} +\eta U_i}{1+\eta}$, where $W_{ij}$ and $U_i$ are independently drawn from $U(0, 1)$. Then we generate $y\sim ~\mbox{Bernoulli}~ \big ( \frac{1}{1+e^{-f^*(\bx)}} \big )$ with the true conditional logit function
$
f^*(\bx)=8x_1+4x_1^2 - 2\cos(\pi x_1/2)+6\sin(\pi(x_2-x_3))-4.
$
Clearly, the first three variables are truly informative.

For both examples, we consider different combinations $(n,p)=(500, 5000)$, $(500, 10000)$, $(500, 50000)$ and $(500,100000)$, and for each case, $\eta=0$ and $\eta=0.5$
are examined. When $\eta=0$, the variables are completely independent, whereas when $\eta=0.5$, correlation structures are added among the variables. Under each setting, the experiment is replicated $100$ times and the averaged performance measures are summarized in Tables \ref{table_example1_eta0}--\ref{table_example2_eta05}, where ``MeanSize'' denotes the averaged number of selected informative variables, ``MaxSize'' denotes the largest number of selected informative variables, ``$X_i$"
refers to the frequency of selecting the corresponding $i$-th covariate variable, and ``C", ``U", ``O" are the frequency of correct-fitting, under-fitting, and over-fitting, respectively.

\begin{table}[!ht]
\scriptsize
\caption{The averaged performance measures of the proposed framework and its competitors in Example 1 with $n=500$ and $\eta=0$.}
\label{table_example1_eta0}
\centering
\begin{tabular}{cccccccccccc}
  \hline
$p$ & Method & $X_1$ & $X_2$ & $X_3$ & $X_4$ & $X_5$ & MaxSize & U & O & C & MeanSize \\
  \hline
5000 & GSLM\_SQ & 1.00 & 1.00 & 1.00 & 1.00 & 1.00 & 6.00 & 0.00 & 0.03 & 0.97 & 5.03 \\
   & GSLM\_QA & 1.00 & 1.00 & 1.00 & 1.00 & 1.00 & 6.00 & 0.00 & 0.04 & 0.96 & 5.04 \\
   & GSLM\_HB & 1.00 & 1.00 & 1.00 & 1.00 & 1.00 & 6.00 & 0.00 & 0.02 & 0.98 & 5.02 \\
   & SIRS-t & 1.00 & 0.99 & 1.00 & 0.96 & 1.00 & 5.00 & 0.04 & 0.00 & 0.96 & 4.95 \\
   & MBKR-t & 1.00 & 1.00 & 1.00 & 0.96 & 1.00 & 5.00 & 0.04 & 0.00 & 0.96 & 4.96 \\
   & DC-t & 1.00 & 0.99 & 1.00 & 0.92 & 1.00 & 5.00 & 0.08 & 0.00 & 0.92 & 4.91 \\
   & Ball-t & 0.96 & 0.99 & 1.00 & 0.87 & 0.99 & 5.00 & 0.15 & 0.00 & 0.85 & 4.81 \\
   & QaSIS-t & 0.93 & 0.97 & 0.96 & 0.85 & 0.97 & 5.00 & 0.26 & 0.00 & 0.74 & 4.68 \\
   \hline
10000 & GSLM\_SQ & 1.00 & 1.00 & 1.00 & 1.00 & 1.00 & 6.00 & 0.00 & 0.02 & 0.98 & 5.02 \\
   & GSLM\_QA & 1.00 & 1.00 & 1.00 & 1.00 & 1.00 & 6.00 & 0.00 & 0.02 & 0.98 & 5.02 \\
   & GSLM\_HB & 1.00 & 1.00 & 1.00 & 1.00 & 1.00 & 6.00 & 0.00 & 0.02 & 0.98 & 5.02 \\
   & SIRS-t & 1.00 & 1.00 & 1.00 & 0.97 & 0.97 & 5.00 & 0.05 & 0.00 & 0.95 & 4.94 \\
   & MBKR-t & 1.00 & 1.00 & 1.00 & 0.98 & 1.00 & 5.00 & 0.02 & 0.00 & 0.98 & 4.98 \\
   & DC-t & 1.00 & 1.00 & 1.00 & 0.97 & 0.97 & 5.00 & 0.05 & 0.00 & 0.95 & 4.94 \\
   & Ball-t & 0.96 & 1.00 & 0.99 & 0.81 & 0.91 & 5.00 & 0.25 & 0.00 & 0.75 & 4.67 \\
   & QaSIS-t & 0.91 & 0.92 & 0.90 & 0.83 & 0.91 & 5.00 & 0.40 & 0.00 & 0.60 & 4.47 \\
   \hline
50000 & GSLM\_SQ & 1.00 & 1.00 & 1.00 & 1.00 & 1.00 & 6.00 & 0.00 & 0.13 & 0.87 & 5.13 \\
   & GSLM\_QA & 1.00 & 1.00 & 1.00 & 1.00 & 1.00 & 6.00 & 0.00 & 0.16 & 0.84 & 5.16 \\
   & GSLM\_HB & 1.00 & 1.00 & 1.00 & 1.00 & 1.00 & 6.00 & 0.00 & 0.16 & 0.84 & 5.16 \\
   & SIRS-t & 1.00 & 0.97 & 0.99 & 0.91 & 0.97 & 5.00 & 0.11 & 0.00 & 0.89 & 4.84 \\
   & MBKR-t & 1.00 & 0.99 & 0.99 & 0.93 & 0.99 & 5.00 & 0.07 & 0.00 & 0.93 & 4.90 \\
   & DC-t & 1.00 & 0.99 & 1.00 & 0.92 & 0.99 & 5.00 & 0.09 & 0.00 & 0.91 & 4.90 \\
   & Ball-t & 0.96 & 0.90 & 0.91 & 0.72 & 0.81 & 5.00 & 0.48 & 0.00 & 0.52 & 4.30 \\
   & QaSIS-t & 0.88 & 0.84 & 0.82 & 0.65 & 0.70 & 5.00 & 0.69 & 0.00 & 0.31 & 3.89 \\
   \hline
100000 & GSLM\_SQ & 1.00 & 1.00 & 1.00 & 1.00 & 1.00 & 9.00 & 0.00 & 0.22 & 0.78 & 5.27 \\
   & GSLM\_QA & 1.00 & 1.00 & 1.00 & 1.00 & 1.00 & 9.00 & 0.00 & 0.25 & 0.75 & 5.33 \\
   & GSLM\_HB & 1.00 & 1.00 & 1.00 & 1.00 & 1.00 & 9.00 & 0.00 & 0.22 & 0.78 & 5.29 \\
   & SIRS-t & 1.00 & 0.98 & 0.98 & 0.90 & 0.97 & 5.00 & 0.12 & 0.00 & 0.88 & 4.83 \\
   & MBKR-t & 1.00 & 0.99 & 0.99 & 0.95 & 1.00 & 5.00 & 0.07 & 0.00 & 0.93 & 4.93 \\
   & DC-t & 1.00 & 1.00 & 1.00 & 0.94 & 1.00 & 5.00 & 0.06 & 0.00 & 0.94 & 4.94 \\
   & Ball-t & 0.92 & 0.94 & 0.93 & 0.66 & 0.82 & 5.00 & 0.52 & 0.00 & 0.48 & 4.27 \\
   & QaSIS-t & 0.79 & 0.74 & 0.82 & 0.51 & 0.66 & 5.00 & 0.82 & 0.00 & 0.18 & 3.52\\
   \hline
\end{tabular}
\end{table}

 \begin{table}[!ht]
 \scriptsize
\caption{The averaged performance measures of the proposed framework and its competitors in Example 1 with $n=500$ and $\eta=0.5$.}
\label{table_example1_eta05}
\centering
\begin{tabular}{cccccccccccc}
  \hline
$p$ & Method & $X_1$ & $X_2$ & $X_3$ & $X_4$ & $X_5$ & MaxSize & U & O & C & MeanSize \\
  \hline
5000 & GSLM\_SQ & 1.00 & 1.00 & 1.00 & 0.99 & 1.00 & 7.00 & 0.01 & 0.03 & 0.96 & 5.03 \\
   & GSLM\_QA & 1.00 & 1.00 & 1.00 & 1.00 & 1.00 & 7.00 & 0.00 & 0.03 & 0.97 & 5.04 \\
   & GSLM\_HB & 1.00 & 1.00 & 1.00 & 1.00 & 1.00 & 7.00 & 0.00 & 0.03 & 0.97 & 5.04 \\
   & SIRS-t & 0.99 & 0.00 & 0.96 & 0.58 & 0.98 & 4.00 & 1.00 & 0.00 & 0.00 & 3.51 \\
   & MBKR-t & 1.00 & 0.00 & 0.98 & 0.72 & 0.99 & 4.00 & 1.00 & 0.00 & 0.00 & 3.69 \\
   & DC-t & 0.98 & 0.00 & 0.99 & 0.62 & 0.98 & 4.00 & 1.00 & 0.00 & 0.00 & 3.57 \\
   & Ball-t & 0.92 & 0.00 & 0.94 & 0.43 & 0.95 & 4.00 & 1.00 & 0.00 & 0.00 & 3.24 \\
   & QaSIS-t & 0.79 & 0.00 & 0.94 & 0.48 & 0.96 & 4.00 & 1.00 & 0.00 & 0.00 & 3.17 \\
   \hline
10000 & GSLM\_SQ & 1.00 & 1.00 & 1.00 & 0.98 & 1.00 & 7.00 & 0.02 & 0.04 & 0.94 & 5.03 \\
   & GSLM\_QA & 1.00 & 1.00 & 1.00 & 0.99 & 1.00 & 7.00 & 0.01 & 0.04 & 0.95 & 5.04 \\
   & GSLM\_HB & 1.00 & 1.00 & 1.00 & 0.98 & 1.00 & 7.00 & 0.02 & 0.04 & 0.94 & 5.03 \\
   & SIRS-t & 1.00 & 0.00 & 0.99 & 0.66 & 1.00 & 4.00 & 1.00 & 0.00 & 0.00 & 3.65 \\
   & MBKR-t & 1.00 & 0.00 & 1.00 & 0.72 & 1.00 & 4.00 & 1.00 & 0.00 & 0.00 & 3.72 \\
   & DC-t & 1.00 & 0.00 & 1.00 & 0.64 & 1.00 & 4.00 & 1.00 & 0.00 & 0.00 & 3.64 \\
   & Ball-t & 0.97 & 0.00 & 0.94 & 0.47 & 0.96 & 4.00 & 1.00 & 0.00 & 0.00 & 3.34 \\
   & QaSIS-t & 0.83 & 0.00 & 0.86 & 0.52 & 0.84 & 4.00 & 1.00 & 0.00 & 0.00 & 3.05 \\
   \hline
50000 & GSLM\_SQ & 1.00 & 1.00 & 1.00 & 0.99 & 1.00 & 6.00 & 0.01 & 0.05 & 0.94 & 5.04 \\
   & GSLM\_QA & 1.00 & 1.00 & 1.00 & 0.99 & 1.00 & 7.00 & 0.01 & 0.11 & 0.88 & 5.11 \\
   & GSLM\_HB & 1.00 & 1.00 & 1.00 & 0.99 & 1.00 & 6.00 & 0.01 & 0.04 & 0.95 & 5.03 \\
   & SIRS-t & 0.92 & 0.00 & 0.96 & 0.39 & 0.99 & 4.00 & 1.00 & 0.00 & 0.00 & 3.26 \\
   & MBKR-t & 0.94 & 0.00 & 0.97 & 0.51 & 1.00 & 4.00 & 1.00 & 0.00 & 0.00 & 3.41 \\
   & DC-t & 0.93 & 0.00 & 0.98 & 0.40 & 0.99 & 4.00 & 1.00 & 0.00 & 0.00 & 3.30 \\
   & Ball-t & 0.85 & 0.00 & 0.90 & 0.20 & 0.89 & 4.00 & 1.00 & 0.00 & 0.00 & 2.84 \\
   & QaSIS-t & 0.57 & 0.00 & 0.70 & 0.28 & 0.73 & 4.00 & 1.00 & 0.00 & 0.00 & 2.27 \\
   \hline
100000 & GSLM\_SQ & 1.00 & 1.00 & 1.00 & 0.97 & 1.00 & 7.00 & 0.03 & 0.08 & 0.89 & 5.06 \\
   & GSLM\_QA & 1.00 & 1.00 & 1.00 & 0.96 & 1.00 & 7.00 & 0.04 & 0.17 & 0.79 & 5.16 \\
   & GSLM\_HB & 1.00 & 1.00 & 1.00 & 0.99 & 1.00 & 7.00 & 0.01 & 0.09 & 0.90 & 5.10 \\
   & SIRS-t & 0.96 & 0.00 & 0.96 & 0.39 & 0.95 & 4.00 & 1.00 & 0.00 & 0.00 & 3.26 \\
   & MBKR-t & 0.94 & 0.00 & 0.95 & 0.45 & 0.98 & 4.00 & 1.00 & 0.00 & 0.00 & 3.32 \\
   & DC-t & 0.95 & 0.00 & 0.97 & 0.41 & 0.95 & 4.00 & 1.00 & 0.00 & 0.00 & 3.28 \\
   & Ball-t & 0.85 & 0.00 & 0.84 & 0.16 & 0.87 & 4.00 & 1.00 & 0.00 & 0.00 & 2.72 \\
   & QaSIS-t & 0.53 & 0.00 & 0.63 & 0.25 & 0.76 & 4.00 & 1.00 & 0.00 & 0.00 & 2.17 \\
   \hline
\end{tabular}
\end{table}

\begin{table}[!ht]
\scriptsize
\caption{The averaged performance measures of the proposed framework and its competitors in Example 2 with $n=500$ and $\eta=0$}
\label{table_example2_eta0}
\centering
\begin{tabular}{cccccccccc}
  \hline
$p$ & Method & $X_1$ & $X_2$ & $X_3$ & MaxSize & U & O & C & MeanSize \\
  \hline
5000 & GSLM-SVM & 1.00 & 0.95 & 0.97 & 5.00 & 0.07 & 0.20 & 0.73 & 3.18 \\
   & GSLM-LOG & 1.00 & 0.93 & 0.95 & 5.00 & 0.11 & 0.05 & 0.84 & 2.94 \\
   & SIRS-t & 0.99 & 0.60 & 0.67 & 3.00 & 0.49 & 0.00 & 0.51 & 2.26 \\
   & MBKR-t & 0.99 & 0.62 & 0.68 & 3.00 & 0.47 & 0.00 & 0.53 & 2.29 \\
   & DC-t & 0.98 & 0.61 & 0.68 & 3.00 & 0.48 & 0.00 & 0.52 & 2.27 \\
   & MVxy-t & 0.98 & 0.60 & 0.67 & 3.00 & 0.49 & 0.00 & 0.51 & 2.25 \\
   & Kol. Filter-t & 0.94 & 0.51 & 0.59 & 3.00 & 0.66 & 0.00 & 0.34 & 2.04 \\
   \hline
10000 & GSLM-SVM & 1.00 & 0.96 & 0.96 & 6.00 & 0.07 & 0.35 & 0.58 & 3.38 \\
   & GSLM-LOG & 1.00 & 0.96 & 0.94 & 4.00 & 0.09 & 0.07 & 0.84 & 2.97 \\
   & SIRS-t & 1.00 & 0.65 & 0.59 & 3.00 & 0.54 & 0.00 & 0.46 & 2.24 \\
   & MBKR-t & 1.00 & 0.63 & 0.58 & 3.00 & 0.53 & 0.00 & 0.47 & 2.21 \\
   & DC-t & 1.00 & 0.65 & 0.65 & 3.00 & 0.51 & 0.00 & 0.49 & 2.30 \\
   & MVxy-t & 1.00 & 0.66 & 0.63 & 3.00 & 0.52 & 0.00 & 0.48 & 2.29 \\
   & Kol. Filter-t & 0.98 & 0.57 & 0.45 & 3.00 & 0.71 & 0.00 & 0.29 & 2.00 \\
   \hline
50000 & GSLM-SVM & 1.00 & 0.93 & 0.98 & 10.00 & 0.08 & 0.15 & 0.77 & 3.50 \\
   & GSLM-LOG & 1.00 & 0.90 & 0.92 & 6.00 & 0.13 & 0.24 & 0.63 & 3.14 \\
   & SIRS-t & 0.98 & 0.42 & 0.45 & 3.00 & 0.82 & 0.00 & 0.18 & 1.85 \\
   & MBKR-t & 0.98 & 0.44 & 0.44 & 3.00 & 0.81 & 0.00 & 0.19 & 1.86 \\
   & DC-t & 0.98 & 0.45 & 0.47 & 3.00 & 0.78 & 0.00 & 0.22 & 1.90 \\
   & MVxy-t & 0.98 & 0.47 & 0.48 & 3.00 & 0.78 & 0.00 & 0.22 & 1.93 \\
   & Kol. Filter-t & 0.90 & 0.31 & 0.35 & 3.00 & 0.92 & 0.00 & 0.08 & 1.56 \\
   \hline
100000 & GSLM-SVM & 1.00 & 0.89 & 0.95 & 15.00 & 0.13 & 0.14 & 0.73 & 3.47 \\
   & GSLM-LOG & 1.00 & 0.84 & 0.97 & 6.00 & 0.18 & 0.32 & 0.50 & 3.28 \\
   & SIRS-t & 0.99 & 0.36 & 0.42 & 3.00 & 0.82 & 0.00 & 0.18 & 1.77 \\
   & MBKR-t & 0.99 & 0.32 & 0.38 & 3.00 & 0.87 & 0.00 & 0.13 & 1.69 \\
   & DC-t & 0.99 & 0.36 & 0.44 & 3.00 & 0.82 & 0.00 & 0.18 & 1.79 \\
   & MVxy-t & 0.99 & 0.37 & 0.44 & 3.00 & 0.82 & 0.00 & 0.18 & 1.80 \\
   & Kol. Filter-t & 0.85 & 0.27 & 0.29 & 3.00 & 0.97 & 0.00 & 0.03 & 1.41 \\
   \hline
\end{tabular}
\end{table}

\begin{table}[!h]
\scriptsize
\caption{The averaged performance measures of the proposed framework and its competitors in Example 2 with $n=500$ and $\eta=0.5$.}
\label{table_example2_eta05}
\centering
\begin{tabular}{cccccccccc}
  \hline
$p$ & Method & $X_1$ & $X_2$ & $X_3$& MaxSize & U & O & C & MeanSize \\
 \hline
5000 & GSLM-SVM & 0.96 & 1.00 & 1.00 & 8.00 & 0.04 & 0.33 & 0.63 & 3.58 \\
   & GSLM-LOG & 0.95 & 1.00 & 1.00 & 9.00 & 0.05 & 0.17 & 0.78 & 3.20 \\
   & SIRS-t & 0.55 & 0.95 & 0.16 & 3.00 & 0.90 & 0.00 & 0.10 & 1.66 \\
   & MBKR-t & 0.53 & 0.95 & 0.18 & 3.00 & 0.89 & 0.00 & 0.11 & 1.66 \\
   & DC-t & 0.54 & 0.94 & 0.18 & 3.00 & 0.89 & 0.00 & 0.11 & 1.66 \\
   & MVxy-t & 0.55 & 0.94 & 0.19 & 3.00 & 0.88 & 0.00 & 0.12 & 1.68 \\
   & Kol. Filter-t & 0.36 & 0.83 & 0.17 & 3.00 & 0.97 & 0.00 & 0.03 & 1.36 \\
   \hline
10000 & GSLM-SVM & 0.95 & 0.99 & 1.00 & 10.00 & 0.06 & 0.28 & 0.66 & 3.69 \\
   & GSLM-LOG & 0.94 & 0.99 & 1.00 & 6.00 & 0.07 & 0.27 & 0.66 & 3.28 \\
   & SIRS-t & 0.63 & 0.94 & 0.08 & 3.00 & 0.94 & 0.00 & 0.06 & 1.65 \\
   & MBKR-t & 0.63 & 0.95 & 0.07 & 3.00 & 0.95 & 0.00 & 0.05 & 1.65 \\
   & DC-t & 0.63 & 0.93 & 0.09 & 3.00 & 0.93 & 0.00 & 0.07 & 1.65 \\
   & MVxy-t & 0.61 & 0.93 & 0.10 & 3.00 & 0.92 & 0.00 & 0.08 & 1.64 \\
   & Kol. Filter-t & 0.37 & 0.73 & 0.10 & 3.00 & 0.98 & 0.00 & 0.02 & 1.20 \\
   \hline
50000 & GSLM-SVM & 0.92 & 1.00 & 1.00 & 27.00 & 0.08 & 0.37 & 0.55 & 4.68 \\
   & GSLM-LOG & 0.88 & 0.99 & 0.99 & 43.00 & 0.14 & 0.41 & 0.45 & 4.40 \\
   & SIRS-t & 0.40 & 0.95 & 0.04 & 3.00 & 0.99 & 0.00 & 0.01 & 1.39 \\
   & MBKR-t & 0.40 & 0.93 & 0.07 & 3.00 & 0.97 & 0.00 & 0.03 & 1.40 \\
   & DC-t & 0.40 & 0.90 & 0.07 & 3.00 & 0.98 & 0.00 & 0.02 & 1.37 \\
   & MVxy-t & 0.39 & 0.90 & 0.07 & 3.00 & 0.98 & 0.00 & 0.02 & 1.36 \\
   & Kol. Filter-t & 0.28 & 0.61 & 0.06 & 2.00 & 1.00 & 0.00 & 0.00 & 0.96 \\
   \hline
100000 & GSLM-SVM & 0.86 & 0.99 & 1.00 & 56.00 & 0.15 & 0.45 & 0.40 & 4.14 \\
   & GSLM-LOG & 0.84 & 0.99 & 0.97 & 14.00 & 0.19 & 0.31 & 0.50 & 4.33 \\
   & SIRS-t & 0.44 & 0.92 & 0.02 & 3.00 & 0.98 & 0.00 & 0.02 & 1.38 \\
   & MBKR-t & 0.43 & 0.91 & 0.02 & 3.00 & 0.98 & 0.00 & 0.02 & 1.36 \\
   & DC-t & 0.43 & 0.89 & 0.04 & 3.00 & 0.97 & 0.00 & 0.03 & 1.36 \\
   & MVxy-t & 0.41 & 0.89 & 0.04 & 3.00 & 0.97 & 0.00 & 0.03 & 1.34 \\
   & Kol. Filter-t & 0.23 & 0.62 & 0.05 & 3.00 & 1.00 & 0.00 & 0.00 & 0.91 \\
   \hline
\end{tabular}
\end{table}

It is evident that the proposed framework outperforms all the competitors in  the both examples. In Example 1, GSLM-SQ, GSLM-QA and GSLM-HB are able to exactly identify all the truly informative variables in most replications. Yet,  all the other competitors tend to miss some truly informative variables.  In Example 2,  GSLM-SVM and GSLM-LOG are also able to identify all the truly informative variables acting on the true conditional logit function with high probability, but  all the other competitors  tend to underfit by missing some important variables. Furthermore, when the correlation structure with $\eta=0.5$ is considered, identifying the truly informative variables becomes more difficult, yet the proposed framework still outperforms the other competitors in most scenarios.

More specifically, the existing methods have included almost all the informative variables as $\eta=0$, but they tend to miss some truly important variables as $\eta=0.5$ even by keeping the first $[n/\log n]$ variables. It is worthy pointing out that the  proposed framework is computationally efficient, which is clearly demonstrated by the computing times given in Table \ref{table_time_example1} based on a
computation machine with eight cores Intel Xeon E5-2695 CPU and 16GB memory.

\begin{table}[!ht]
\scriptsize
  \centering
  \caption{Comparison of all the methods in terms of  averaged run-time (in seconds) in Examples 1 and 2.}
    \label{table_time_example1}%

    \begin{tabular}{cccccccccc}
    \hline
      & $p$    & GSLM-SQ & GSLM-QA & GSLM-HB & MBKR  & SIRS  &DC   & Ball &  QaSIS\\
          \hline
Example 1 &    5000  & 6.2 & 7.2 & 6.9 & 655.4 & 21.2 & 114.4 & 9.4 & 11.9 \\
&  10000 &9.4 & 9.8 & 9.0 & 1106.9 & 41.8 & 226.4 &  19.0 & 24.7\\
& 50000 &     43.0 &  41.4  &    36.4     & 5656.6& 200.8 & 1059.0 & 91.5 & 114.0 \\
&100000 & 108.8 & 97.9 & 93.9 & 11494.1 & 468.4 & 2612.0 & 213.7 & 274.8 \\
\hline
 &     & GSLM-SVM & GSLM-LOG & &   MBKR  & SIRS  & DC  & MV-SIS & Kol. Filter  \\
          \hline
Example 2 &    5000  & 5.9 & 9.2 &  & 266.7 & 18.6 & 116.2 & 71.6 &54.6 \\
&  10000 &8.8 & 11.9 & &523.0 & 36.0 & 231.3 &  139.5& 105.1\\
& 50000 &     35.7 &  38.1 &        & 2783.0& 179.9 & 1106.3 & 691.5 & 515.2 \\
&100000 & 98.7 & 101.3 &  & 6443.3 & 405.2 &  2607.9& 1542.8 &1146.2\\
 \hline
    \end{tabular}%
\end{table}%

\subsection{Application to interaction selection}
In this part, the application of the proposed framework to interaction selection is considered.
Specifically,  we consider regression  with  the squared loss, the check loss with $\tau=0.5$ and the Huber loss, and compare the performance with four competitors, including the regularized interaction selection method (RAMP, \cite{Hao2018}), the interaction pursuit with distance correlation (IPDC, \cite{Kong2017}), and the forward selection methods (iFort and iForm, \cite{Hao2014}). For IPDC, we also report the truncated results and denote it as  IPDC-t.
Note that  the computational cost of the existing nonparametric interaction selection methods \citep{ Radchenko2010, Lin2006,  Dong2021} is very expensive, and thus they are not included in the numerical study where large dimensions are considered.

The following  simulated example is examined under various scenarios.

{\noindent\bf Example 3}:
The generating scheme is the same as Example 1 except that the response $y_i$ is generated as
$
y_i=
2\left( f(x_{i1})
-f(x_{i2})
+ f(x_{i3})
- f(x_{i4})\right)
+ 5\pi \left( g(x_{i1},x_{i2})
- g(x_{i2},x_{i3})
+-  g(x_{i3},x_{i4})\right)
+\epsilon_i,
$
where $f(u)=\exp(u)$, $g(u,v)=\cos^2(\pi uv)$, and $\epsilon_i$'s are independently drawn from $N(0,1)$. Clearly, the first four variables are truly informative and the informative interaction terms are ($X_1X_2$, $X_2X_3$, $X_3X_4$).

For Example 3, we consider the same scenarios as those of Section \ref{sec:app1}, and the averaged performance measures are summarized in Tables \ref{table_example3_eta0}--\ref{table_example3_eta05},
where ``$S_M$'' denotes the frequency of covering all the  four main effects, ``NumMain'' denotes the average number of selected main effects, ``$X_iX_j$" refers to the frequency of selecting the corresponding interaction effects between the $i$-th and $j$-th covariates, ``NumInter'' denotes the averaged number of selected interaction effects, ``MaxInter'' denotes the maximum number of selected interaction effects, and ``$C_I$", ``$U_I$", ``$O_I$" are the frequency of correct-fitting, under-fitting, and over-fitting in terms of  interaction effects, respectively.

\begin{table}[!ht]
\scriptsize
   \caption{The averaged performance measures of the proposed framework and its competitors in Example 3 with $n=500$ and $\eta=0$. Note that IPDC selects $[n/\log n]$ main effects and $[n/ \log n]$ interaction terms.}
   \label{table_example3_eta0}
\centering

\begin{tabular}{cccccccccccc}
  \hline
$p$ & Method & $S_M$ & NumMain & $X_1X_2$ & $X_2X_3$ & $X_3X_4$ & $U_I$ & $O_I$ & $C_I$ & NumInter & MaxInter \\
  \hline
5000 & GSLM-SQ & 1.00 & 4.11 & 1.00 & 1.00 & 1.00 & 0.00 & 0.00 & 1.00 & 3.00 & 3.00 \\
   & GSLM-QA & 0.99 & 4.17 & 0.96 & 0.90 & 0.95 & 0.13 & 0.01 & 0.86 & 2.82 & 4.00 \\
   & GSLM-HB & 1.00 & 4.11 & 1.00 & 1.00 & 1.00 & 0.00 & 0.00 & 1.00 & 3.00 & 3.00 \\
   & RAMP & 1.00 & 4.05 & 0.00 & 0.00 & 0.00 & 1.00 & 0.00 & 0.00 & 0.00 & 0.00 \\
   & iFort & 1.00 & 4.00 & 0.00 & 0.00 & 0.00 & 1.00 & 0.00 & 0.00 & 0.00 & 0.00 \\
   & iForm & 1.00 & 4.00 & 0.00 & 0.00 & 0.00 & 1.00 & 0.00 & 0.00 & 0.00 & 0.00 \\
   & IPDC & 1.00 & 81.00 & 1.00 & 0.75 & 1.00 & 0.25 & 0.75 & 0.00 & 81.00 & 81.00 \\
   & IPDC-t & 0.93 & 3.88 & 0.00 & 0.00 & 0.00 & 1.00 & 0.00 & 0.00 & 0.00 & 0.00 \\
   \hline
10000 & GSLM-SQ & 1.00 & 4.07 & 1.00 & 1.00 & 1.00 & 0.00 & 0.00 & 1.00 & 3.00 & 3.00 \\
   & GSLM-QA & 1.00 & 4.12 & 0.98 & 0.96 & 0.99 & 0.05 & 0.00 & 0.95 & 2.99 & 6.00 \\
   & GSLM-HB & 1.00 & 4.13 & 1.00 & 1.00 & 1.00 & 0.00 & 0.00 & 1.00 & 3.00 & 3.00 \\
   & RAMP & 1.00 & 4.07 & 0.00 & 0.00 & 0.00 & 1.00 & 0.00 & 0.00 & 0.00 & 0.00 \\
   & iFort & 1.00 & 4.00 & 0.00 & 0.00 & 0.00 & 1.00 & 0.00 & 0.00 & 0.00 & 0.00 \\
   & iForm & 1.00 & 4.00 & 0.00 & 0.00 & 0.00 & 1.00 & 0.00 & 0.00 & 0.00 & 0.00 \\
   & IPDC & 1.00 & 81.00 & 1.00 & 0.85 & 0.99 & 0.16 & 0.84 & 0.00 & 81.00 & 81.00 \\
   & IPDC-t & 0.93 & 3.91 & 0.00 & 0.00 & 0.00 & 1.00 & 0.00 & 0.00 & 0.00 & 0.00 \\
   \hline
50000 & GSLM-SQ & 1.00 & 4.15 & 1.00 & 1.00 & 1.00 & 0.00 & 0.00 & 1.00 & 3.00 & 3.00 \\
   & GSLM-QA & 1.00 & 4.24 & 0.95 & 0.88 & 0.93 & 0.13 & 0.01 & 0.86 & 2.79 & 5.00 \\
   & GSLM-HB & 1.00 & 4.13 & 1.00 & 1.00 & 1.00 & 0.00 & 0.00 & 1.00 & 3.00 & 3.00 \\
   & RAMP & 1.00 & 4.06 & 0.00 & 0.00 & 0.00 & 1.00 & 0.00 & 0.00 & 0.00 & 0.00 \\
   & iFort & 1.00 & 4.00 & 0.00 & 0.00 & 0.00 & 1.00 & 0.00 & 0.00 & 0.00 & 0.00 \\
   & iForm & 1.00 & 4.00 & 0.00 & 0.00 & 0.00 & 1.00 & 0.00 & 0.00 & 0.00 & 0.00 \\
   & IPDC & 1.00 & 81.00 & 1.00 & 0.81 & 1.00 & 0.19 & 0.81 & 0.00 & 81.00 & 81.00 \\
   & IPDC-t & 0.90 & 3.88 & 0.00 & 0.00 & 0.00 & 1.00 & 0.00 & 0.00 & 0.00 & 0.00 \\
   \hline
100000 & GSLM-SQ & 0.98 & 4.12 & 0.98 & 0.99 & 1.00 & 0.02 & 0.00 & 0.98 & 2.98 & 3.00 \\
   & GSLM-QA & 0.99 & 4.14 & 0.98 & 0.96 & 0.98 & 0.06 & 0.02 & 0.92 & 2.97 & 7.00 \\
   & GSLM-HB & 0.98 & 4.10 & 0.98 & 0.99 & 1.00 & 0.02 & 0.00 & 0.98 & 2.98 & 3.00 \\
   & RAMP & 1.00 & 4.03 & 0.00 & 0.00 & 0.00 & 1.00 & 0.00 & 0.00 & 0.00 & 0.00 \\
   & iFort & 1.00 & 4.00 & 0.00 & 0.00 & 0.00 & 1.00 & 0.00 & 0.00 & 0.00 & 0.00 \\
   & iForm & 1.00 & 4.00 & 0.00 & 0.00 & 0.00 & 1.00 & 0.00 & 0.00 & 0.00 & 0.00 \\
   & IPDC & 1.00 & 81.00 & 1.00 & 0.76 & 0.98 & 0.26 & 0.74 & 0.00 & 81.00 & 81.00 \\
   & IPDC-t & 0.78 & 3.70 & 0.00 & 0.00 & 0.00 & 1.00 & 0.00 & 0.00 & 0.00 & 0.00 \\
   \hline
\end{tabular}
\end{table}

\begin{table}[!ht]
\scriptsize
   \caption{The averaged performance measures of the proposed framework and its competitors in   Example 3 with $n=500$ and $
   \eta=0.5$. Note that IPDC selects $[n/\log n]$ main effects and $[n/\log n]$ interaction terms.}
   \label{table_example3_eta05}
\centering
\begin{tabular}{cccccccccccc}
  \hline
$p$ & Method & $S_M$ & NumMain & $X_1X_2$ & $X_2X_3$ & $X_3X_4$ & $U_I$ & $O_I$ & $C_I$ & NumInter & MaxInter \\
\hline
5000 & GSLM-SQ & 1.00 & 4.07 & 1.00 & 1.00 & 1.00 & 0.00 & 0.17 & 0.83 & 3.17 & 4.00 \\
   & GSLM-QA & 1.00 & 4.07 & 1.00 & 0.98 & 1.00 & 0.02 & 0.15 & 0.83 & 3.14 & 5.00 \\
   & GSLM-HB & 1.00 & 4.04 & 1.00 & 1.00 & 1.00 & 0.00 & 0.22 & 0.78 & 3.22 & 4.00 \\
   & RAMP & 1.00 & 4.02 & 0.48 & 0.09 & 0.06 & 1.00 & 0.00 & 0.00 & 0.63 & 2.00 \\
   & iFort & 1.00 & 4.00 & 0.00 & 0.00 & 0.00 & 1.00 & 0.00 & 0.00 & 0.00 & 0.00 \\
   & iForm & 1.00 & 4.00 & 0.14 & 0.00 & 0.00 & 1.00 & 0.00 & 0.00 & 0.14 & 1.00 \\
   & IPDC & 1.00 & 81.00 & 0.87 & 0.35 & 0.91 & 0.79 & 0.21 & 0.00 & 81.00 & 81.00 \\
   & IPDC-t & 0.91 & 3.88 & 0.00 & 0.00 & 0.00 & 1.00 & 0.00 & 0.00 & 0.00 & 0.00 \\
   \hline
10000 & GSLM-SQ & 1.00 & 4.11 & 1.00 & 1.00 & 1.00 & 0.00 & 0.14 & 0.86 & 3.14 & 4.00 \\
   & GSLM-QA & 1.00 & 4.10 & 0.98 & 0.95 & 1.00 & 0.06 & 0.14 & 0.80 & 3.08 & 5.00 \\
   & GSLM-HB & 1.00 & 4.09 & 1.00 & 1.00 & 0.99 & 0.01 & 0.14 & 0.85 & 3.14 & 5.00 \\
   & RAMP & 1.00 & 4.03 & 0.47 & 0.13 & 0.09 & 0.97 & 0.00 & 0.03 & 0.69 & 3.00 \\
   & iFort & 1.00 & 4.00 & 0.00 & 0.00 & 0.00 & 1.00 & 0.00 & 0.00 & 0.00 & 0.00 \\
   & iForm & 1.00 & 4.00 & 0.12 & 0.01 & 0.00 & 1.00 & 0.00 & 0.00 & 0.13 & 1.00 \\
   & IPDC & 1.00 & 81.00 & 0.94 & 0.39 & 0.87 & 0.73 & 0.27 & 0.00 & 81.00 & 81.00 \\
   & IPDC-t & 0.83 & 3.77 & 0.00 & 0.00 & 0.00 & 1.00 & 0.00 & 0.00 & 0.00 & 0.00 \\
   \hline
50000 & GSLM-SQ & 0.99 & 4.12 & 0.99 & 0.99 & 1.00 & 0.01 & 0.13 & 0.86 & 3.14 & 5.00 \\
   & GSLM-QA & 1.00 & 4.21 & 1.00 & 0.94 & 0.97 & 0.08 & 0.14 & 0.78 & 3.06 & 4.00 \\
   & GSLM-HB & 0.99 & 4.12 & 0.99 & 0.99 & 1.00 & 0.01 & 0.15 & 0.84 & 3.16 & 5.00 \\
   & RAMP & 1.00 & 4.03 & 0.33 & 0.10 & 0.05 & 0.99 & 0.00 & 0.01 & 0.48 & 3.00 \\
   & iFort & 1.00 & 4.00 & 0.00 & 0.00 & 0.00 & 1.00 & 0.00 & 0.00 & 0.00 & 0.00 \\
   & iForm & 1.00 & 4.00 & 0.11 & 0.00 & 0.00 & 1.00 & 0.00 & 0.00 & 0.11 & 1.00 \\
   & IPDC & 1.00 & 81.00 & 0.91 & 0.40 & 0.90 & 0.69 & 0.31 & 0.00 & 81.00 & 81.00 \\
   & IPDC-t & 0.60 & 3.45 & 0.00 & 0.00 & 0.00 & 1.00 & 0.00 & 0.00 & 0.00 & 0.00 \\
   \hline
100000 & GSLM-SQ & 1.00 & 4.19 & 1.00 & 0.99 & 1.00 & 0.01 & 0.12 & 0.87 & 3.11 & 4.00 \\
   & GSLM-QA & 1.00 & 4.21 & 0.97 & 0.96 & 0.99 & 0.06 & 0.13 & 0.81 & 3.08 & 5.00 \\
   & GSLM-HB & 0.99 & 4.22 & 0.99 & 0.98 & 0.99 & 0.02 & 0.12 & 0.86 & 3.10 & 4.00 \\
   & RAMP & 1.00 & 4.01 & 0.32 & 0.06 & 0.02 & 1.00 & 0.00 & 0.00 & 0.40 & 2.00 \\
   & iFort & 1.00 & 4.00 & 0.00 & 0.00 & 0.00 & 1.00 & 0.00 & 0.00 & 0.00 & 0.00 \\
   & iForm & 1.00 & 4.00 & 0.06 & 0.00 & 0.00 & 1.00 & 0.00 & 0.00 & 0.06 & 1.00 \\
   & IPDC & 1.00 & 81.00 & 0.86 & 0.50 & 0.89 & 0.66 & 0.34 & 0.00 & 81.00 & 81.00 \\
   & IPDC-t & 0.68 & 3.48 & 0.00 & 0.00 & 0.00 & 1.00 & 0.00 & 0.00 & 0.00 & 0.00 \\
   \hline
\end{tabular}
\end{table}

From Tables \ref{table_example3_eta0}--\ref{table_example3_eta05}, it is clear that  the proposed framework outperforms all its competitors in that it can exactly select all the non-linear interaction effects with high probability, while the other methods tend to under-fitting. When $\eta=0$, the proposed framework is the best performer, followed by  IPDC, which keeps $[n/ \log n]$ interaction effects and still tends to underfiting. However, IPDC-t  fails  in all the scenarios, which implies that the  ranking estimated by IPDC may  not be accurate.  All the other competitors fail to detect the underlying interaction structure largely due to most of them are designed for the parametric case. When the correlation structure with $\eta=0.5$ is considered, identifying the truly interaction terms becomes more difficult, yet the proposed framework still achieves the best performance  in  all the scenarios.

It is interesting to notice that the poor performance of some methods, including RAMP, iFort and iForm, is probably due to the fact that they are designed for the parametric cases and  the marginal linear correlations between the interaction terms and the response in Example 3 is quite weak. We refer to the Supplementary Material for the additional comparison under a parametric setting, where the similar phenomenon can also be observed.

\subsection{Real application to the human breast cancer study}\label{CaseStudy}
In this section, we apply the proposed framework to a real dataset on  the human breast cancer study \citep{ZhangWU2016}, which can be downloaded  at \url{https://www.ncbi.nlm.nih.gov/geo/} with accessing number GSE20194. It consists of 278 patients, whereas 164 of them have positive oestrogen receptor status and the other 114  have negative oestrogen receptor status, and each patient is  characterized by 22283 probs.
A patient has positive oestrogen receptor status if  the receptors for estrogen are detected, which suggests that estrogen may send signals to the cancer cells among  normal breast cells to promote their growth. It has been shown that roughly 80 percent of the patients diagnosed with breast cancers,  have the positive estrogen receptor status. Consequently, the main interest of the study is to identify those genes related with the oestrogen receptor status.

For interpretability, we map the prob IDs to the gene symbol and delete the IDs that cannot be mapped. The map relationship is also provided by \url{https://www.ncbi.nlm.nih.gov/geo/}. Finally, 19820 genes are considered  in our application. Clearly, the response variable in this dataset is binary, and thus we apply all the methods used in Example 2 to identify the informative genes. The genes selected by the proposed framework and the competitors are  reported in Table \ref{table_realdata_gene_select}.

\begin{table}[!htbp]
\tiny
\centering
  \caption{The genes selected by the proposed framework and its competitors in the application to the human breast cancer study.}
  \label{table_realdata_gene_select}
     \begin{tabular}{c|r|cccccccccc}
    \hline
    Method & Number & \multicolumn{10}{c}{Selected Genes.} \\
\hline
GSLM-SVM & 10    & CDH3  & ESR1  & GREB1 & AGR2  & PRKD3 & TNNT1 & NAT1  & HOXA1 & VGLL1 & IRX4 \\
GSLM-LOG & 26    & SCCPDH & SPTLC2 & CDH3  & CA12  & PTPRG & ESR1  & REPS2 & GREB1 & ZIC1  & SCGB1D2 \\
&       & SLC15A1 & SEPT9 & AGR2  & ABCC3 & BLZF1 & PRKD3 & ANXA9 & TNNT1 & NAT1  & HOXA1 \\
&       & VGLL1 & VAV3  & SLC37A1 & MBNL3 & IRX4  & NPAS2 &       &       &       &  \\
SIRS-t & 7     & SLC39A6 & CA12  & ESR1  & AGR2  & GATA3 & TBC1D9 & NAT1  &       &       &   \\
MBKR-t & 4     & ESR1  & GATA3 & TBC1D9 & CA12  &       &       &       &       &       &   \\
MV-SIS-t & 4     & ESR1  & GATA3 & TBC1D9 & CA12  &       &       &       &       &       &   \\
DC-t  & 1     & ESR1  &       &       &       &       &       &       &       &       &   \\
Kol.Filter-t & 1     & ESR1  &       &       &       &       &       &       &       &       &   \\
SIRS & 36    & ESR1  & GATA3 & TBC1D9 & CA12  & NAT1  & SLC39A6 & AGR2 & FOXA1 &GREB1 & MLPH \\
&       & DNAJC12 & VAV3  & C6orf211 & XBP1  & VGLL1 & KDM4B & ANXA9 & CDH3 & DNALI1 & \multicolumn{1}{l}{IL6ST} \\
&       & UGCG  & TFF1  & MKL2  & SCCPDH & EVL   & IGF1R & \multicolumn{1}{l}{TTC39A} & \multicolumn{1}{l}{METRN} & \multicolumn{1}{l}{GFRA1} & \multicolumn{1}{l}{MYB} \\
&       & PBX1  & CERS6 & WWP1  & MCCC2 & IGFBP4 & ABAT  &       &       &       &  \\
MBKR & 36    & ESR1  & GATA3 & TBC1D9 & CA12  & NAT1  & C6orf211 & \multicolumn{1}{l}{SLC39A6} & \multicolumn{1}{l}{FOXA1} & \multicolumn{1}{l}{DNAJC12} & \multicolumn{1}{l}{GREB1} \\
&       & KDM4B & IGF1R & UGCG  & VAV3  & MKL2  & EVL   & \multicolumn{1}{l}{IL6ST} & \multicolumn{1}{l}{ANXA9} & \multicolumn{1}{l}{AGR2} & \multicolumn{1}{l}{ABAT} \\
&       & GFRA1 & TTC39A & MAGED2 & MLPH  & MCCC2 & WWP1  & \multicolumn{1}{l}{XBP1} & \multicolumn{1}{l}{SCCPDH} & \multicolumn{1}{l}{RABEP1} & {CDH3} \\
&       & EGFR  & TFF1  & VGLL1 & DNALI1 & DACH1 & MYB   &       &       &       &  \\
MV-SIS & 36    & ESR1  & GATA3 & TBC1D9 & CA12  & NAT1  & C6orf211 & \multicolumn{1}{l}{SLC39A6} & \multicolumn{1}{l}{DNAJC12} & {FOXA1} & \multicolumn{1}{l}{GREB1} \\
&       & IGF1R & KDM4B & UGCG  & VAV3  & MKL2  & IL6ST & \multicolumn{1}{l}{EVL} & {ANXA9} & \multicolumn{1}{l}{GFRA1} & \multicolumn{1}{l}{ABAT} \\
&       & MCCC2 & AGR2  & MAGED2 & WWP1  & TFF1  & EGFR  & \multicolumn{1}{l}{DNALI1} & \multicolumn{1}{l}{XBP1} & \multicolumn{1}{l}{TTC39A} & \multicolumn{1}{l}{RABEP1} \\
&       & MLPH  & SCCPDH & CDH3  & VGLL1 & DACH1 & COX6C &       &       &       &  \\
DC & 36    & ESR1  & GATA3 & TBC1D9 & CA12  & NAT1  & C6orf211 & \multicolumn{1}{l}{SLC39A6} & \multicolumn{1}{l}{FOXA1} & \multicolumn{1}{l}{AGR2} & \multicolumn{1}{l}{DNAJC12} \\
&       & GREB1 & MLPH  & KDM4B & VAV3  & MKL2  & IL6ST & \multicolumn{1}{l}{EVL} & \multicolumn{1}{l}{IGF1R} & \multicolumn{1}{l}{ANXA9} & \multicolumn{1}{l}{VGLL1} \\
&       & UGCG  & XBP1  & GFRA1 & DNALI1 & TFF1  & TTC39A & \multicolumn{1}{l}{ABAT} & \multicolumn{1}{l}{WWP1} & \multicolumn{1}{l}{CDH3} & \multicolumn{1}{l}{MCCC2} \\
&       & SCCPDH & MAGED2 & RABEP1 & MYB   & METRN & PBX1  &       &       &       &  \\
Kol.Filter & 36    & ESR1  & GATA3 & TBC1D9 & NAT1  & CA12  & C6orf211 & \multicolumn{1}{l}{IL6ST} & \multicolumn{1}{l}{SLC39A6} & \multicolumn{1}{l}{UGCG} & \multicolumn{1}{l}{ANXA9} \\
&       & DNAJC12 & EVL   & GREB1 & TFF1  & ABAT  & FOXA1 & \multicolumn{1}{l}{MKL2} & \multicolumn{1}{l}{VAV3} & \multicolumn{1}{l}{IGF1R} & \multicolumn{1}{l}{KDM4B} \\
&       & MYB   & MLPH  & GFRA1 & MCCC2 & VGLL1 & DNALI1 & \multicolumn{1}{l}{COX6C} & \multicolumn{1}{l}{RARA} & \multicolumn{1}{l}{BTG3} & \multicolumn{1}{l}{SLC44A4} \\
&       & WWP1  & CLSTN2 & XBP1  & EGFR  & AGR2  & SCCPDH &       &       &       &  \\
\hline
    \end{tabular}%
\end{table}%

Clearly, GSLM-SVM selects 10 genes and GSLM-LOG selects 26 important genes while  all the other screening-based methods keep 36 genes as suggested  and their truncated versions select at most 7 genes. It is interesting to point out  that  four genes, including PRKD3, TNNT1, HOXA1 and IRX4, are identified by  GSLM-SVM and GLSM-LOG, but missed by all the other competitors.  More importantly, literature search suggests that  these  genes  have important biological implications. Specifically, PRKD3 functions as an important oncogenic driver in the invasive breast cancer \citep{liu2017PRKD3}; TNNT1 facilitates proliferation of breast cancer cells by promoting the G1/S phase transition \citep{shi2018tnnt1}; HOXA1 upregulation is associated with poor prognosis and tumor progression in the breast cancer \citep{liu2019hoxa1};  \cite{correa2017IRX4} discovers the high levels expression of IRX4 in the breast cancer plasma samples.

\begin{figure}[!h]
\includegraphics[width=0.5\textwidth]{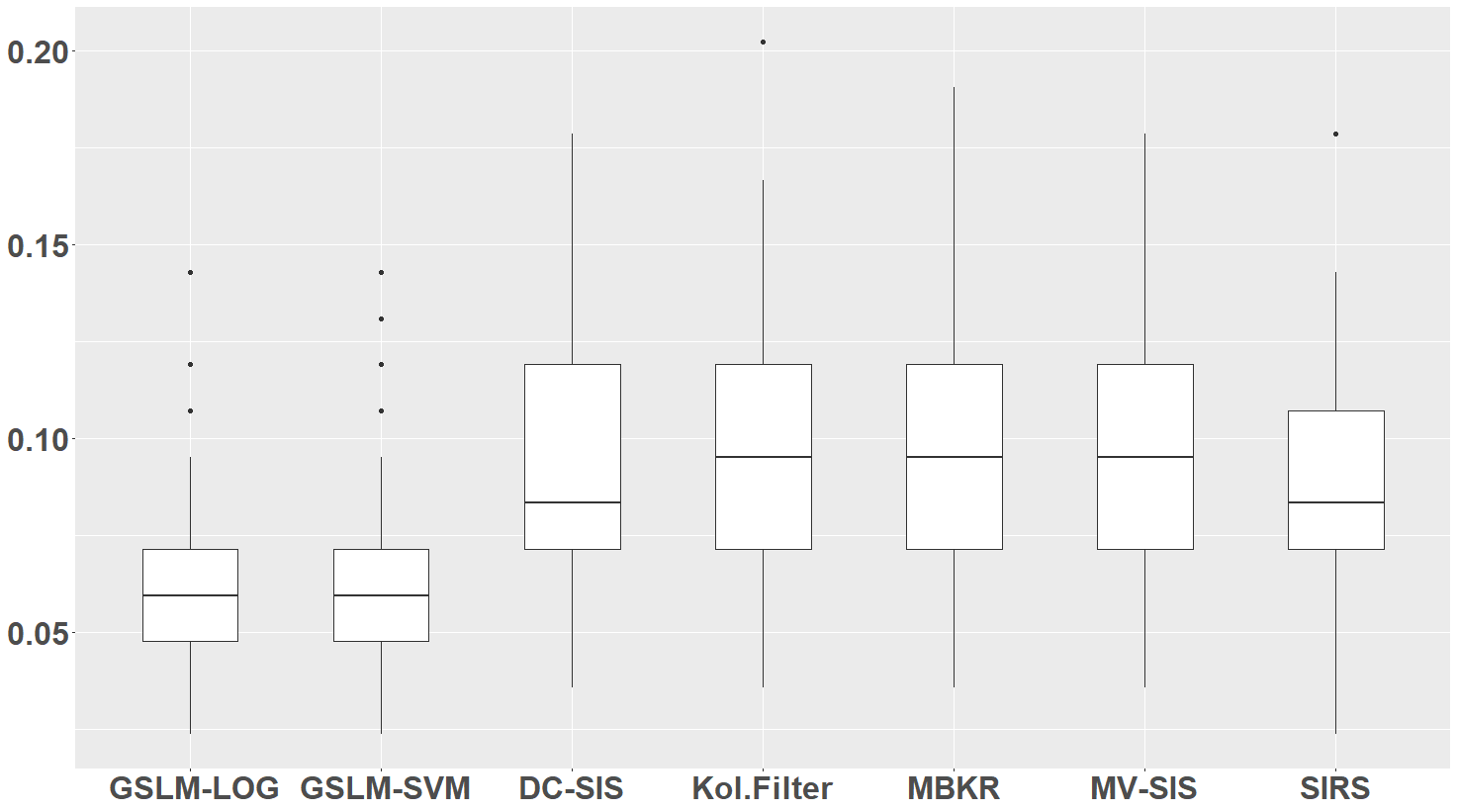} \includegraphics[width=0.5\textwidth]{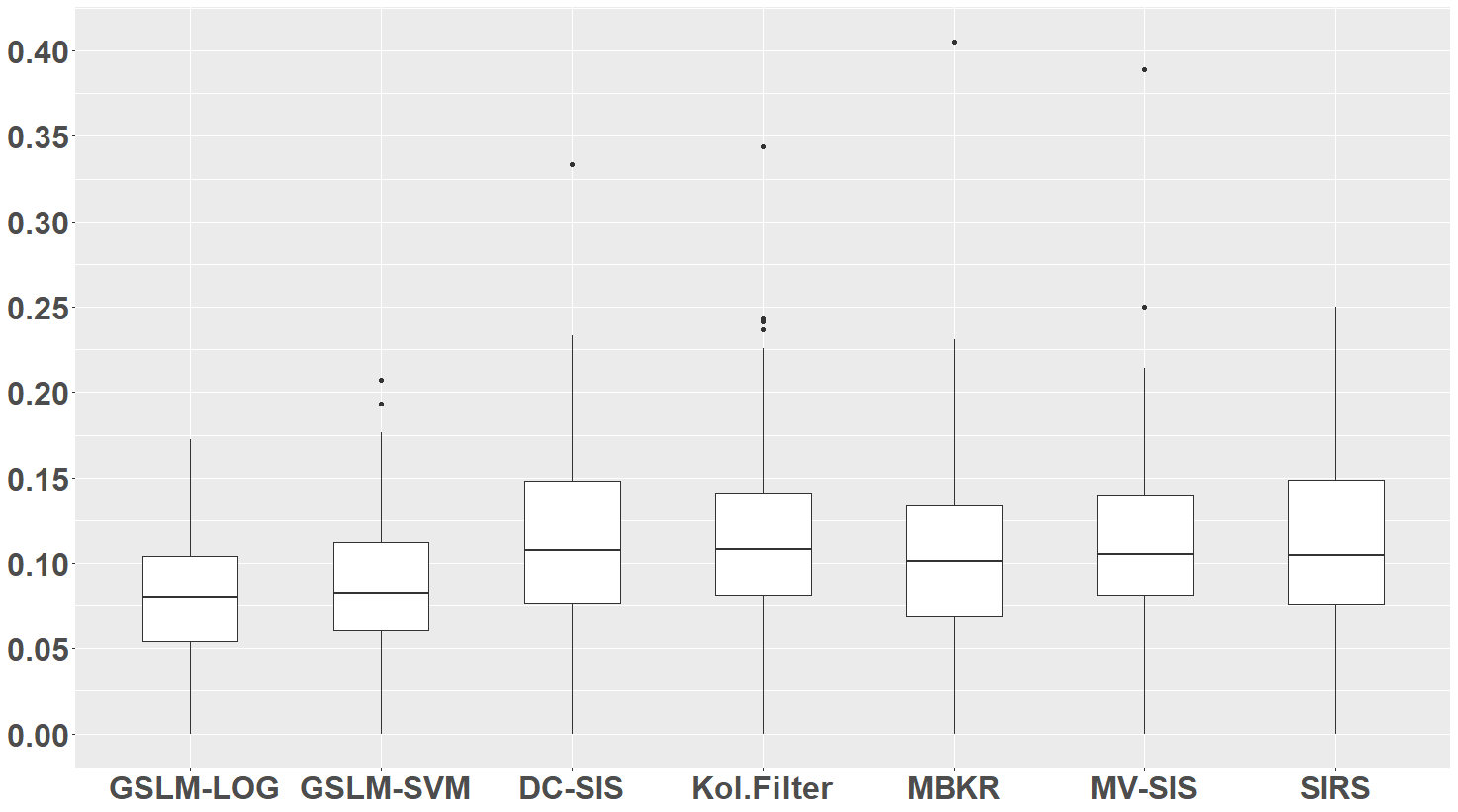} \caption{The boxplots of the testing errors (left panel) and the false negative rates (right panel) of all the methods considered in Section \ref{CaseStudy}}\label{RealComp}
\end{figure}

To support the  superior performance of the proposed framework, we also evaluate the prediction accuracy of the proposed framework and all the screening-based methods  given the their selected genes. Specifically,
we randomly split the dataset with 84 (30$\%$) patients for testing and
the rest for training, and  refit a standard kernel SVM by using the R package \textit{kernlab}. The splitting process is replicated 100 times, and the boxplots of the prediction errors are given in the left panel of Figure \ref{RealComp}.  Since the oestrogen receptor status plays an important role in assisting diagnosis for the breast cancer, it is more severe to miss-classify the patients with positive oestrogen receptor status to be negative. Therefore, we also summarize the false negative rates in the right panel of Figure \ref{RealComp}.

It is clear that  both the averaged testing error and the false negative rate based on the selected sets of GSLM-LOG  are the smallest among all the methods, and followed by GSLM-SVM. Note that the GSLM-LOG selects 26 genes and the GSLM-SVM selects 10 genes while the other screening based method select 36 genes.
 This implies that the proposed method has probably identified some important  genes   missed by the existing methods.

\section{Discussion}\label{sec:dis}
It is known that continuous functions can be well approximated by those functions of the RKHS induced by some universal kernels under the infinity norm. We thus propose a general structure learning framework within the induced RKHS, which can be  used to solve many interesting statistical problems, such as sparse learning, interaction selection, model identification and so on.  The proposed framework is inspired by   the fact that gradient functions can be employed  to define the underlying structures of true target function without model specifications, and  the nice properties of the RKHS facilitate the whole computation of the proposed framework. It is methodologically simple and computationally easy to implement, and can efficiently process large datasets. More importantly, it attains many advantages that it  works for  a general family of loss functions, and admits general dependence structures  with theoretical guarantees under weaker conditions than existing methods. In our future work, we may extend current work to  more complicated cases such as manifold learning and graph estimation.


\section*{Supplementary Material}
Due to the space limit, additional numerical results and all the technical proofs of Theorems \ref{thm1}--\ref{thm:iden} are deferred to  the Supplementary Material.

\section*{Acknowledgment}
Xin He's research is supported in part by NSFC-11901375, Shanghai Pujiang Program 2019PJC051 and the Fundamental Research Funds for the Central Universities. Xingdong Feng's  research is supported in part by  NSFC-11971292 and 11690012, and Program for Innovative Research Team of
SUFE.

\bibliography{bibtex}
\bibliographystyle{plainnat}



 \end{document}